\documentclass[10pt,twocolumn,letterpaper]{article}

\usepackage[pagenumbers]{cvpr}

\usepackage{cuted} 

\usepackage{currfile} 

\usepackage{caption} 
\definecolor{cvprblue}{rgb}{0.21,0.49,0.74}
\definecolor{tableblue}{RGB}{227, 235, 244}
\usepackage[pagebackref,breaklinks,colorlinks,allcolors=cvprblue]{hyperref}
\usepackage{booktabs}
\usepackage{multirow}
\usepackage[table]{xcolor}
\usepackage{graphicx}
\usepackage{bm}
\usepackage{amsmath,amssymb,booktabs,tabularx,array,xcolor,graphicx,tikz,listings}

\def\confName{CVPR}
\def\confYear{2027}

\title{Decoupled Adaptation, Joint Alteration: \\ Instruction-Guided Audio-Video Editing Made Simple}

\author{
  Yujia Hu \qquad Songhua Liu\thanks{Corresponding author.}\\
  Shanghai Jiao Tong University
}

\makeatletter
\apptocmd{\@maketitle}{%
    \par
    \noindent
    \begin{minipage}{\textwidth}
        \centering
\includegraphics[width=\linewidth]{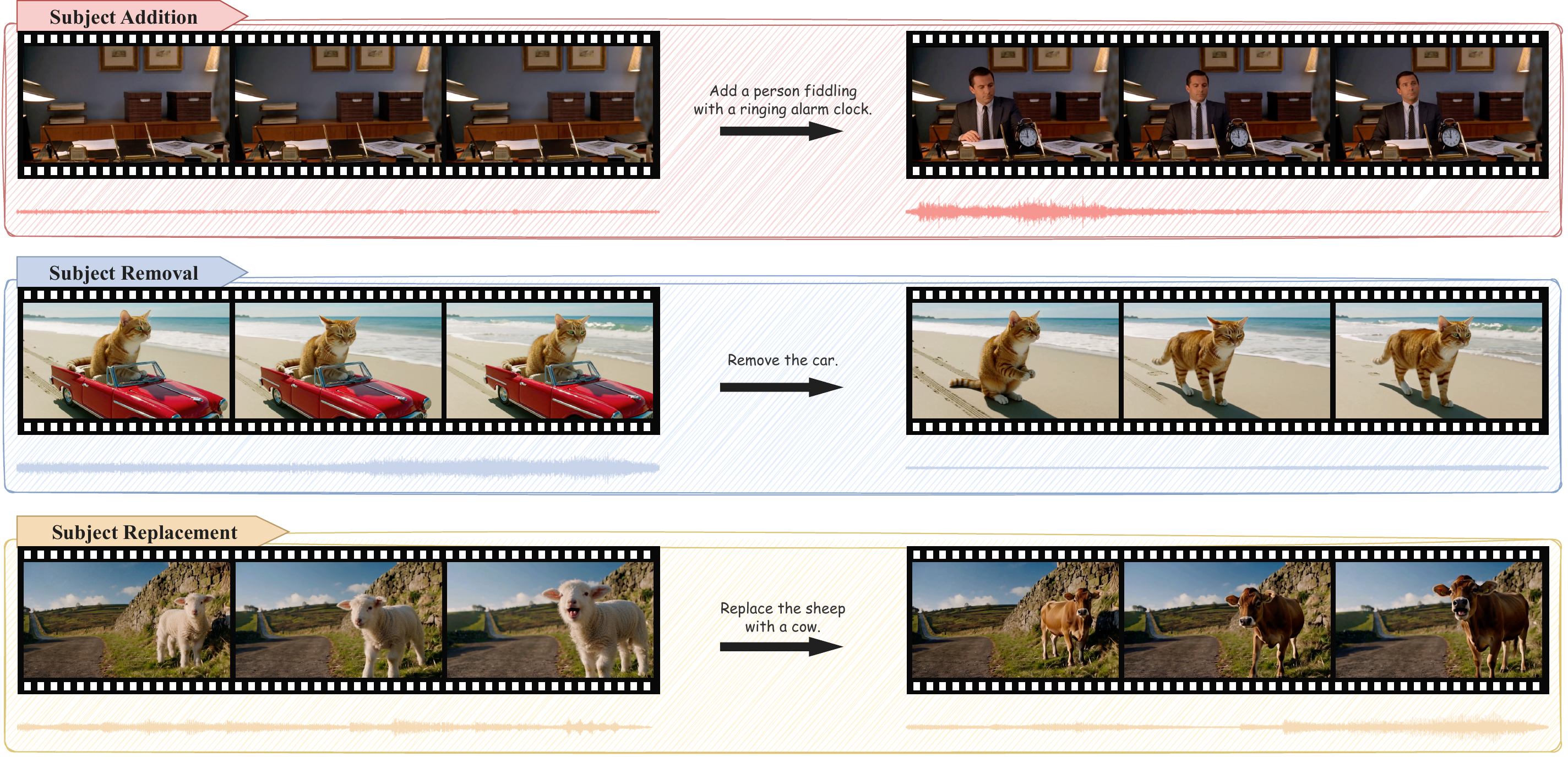}
\captionof{figure}{Results of our proposed method. Given a source video and an editing instruction, DAJA coordinates visual and acoustic modifications while preserving unrelated content and maintaining audio-visual consistency.
}
\label{fig:teaser}
    \end{minipage}
    \par
    \vspace{10pt}
}{}{\PackageError{teaser}{Failed to patch title}{}}
\makeatother
\begin{document}
\maketitle
\begin{abstract}
Recent advances in audio-video generation have demonstrated strong capabilities in jointly modeling visual content and sound. 
However, instruction-guided editing of existing audio-video content remains challenging, as it requires coordinated modifications to both modalities while preserving unrelated content and audio-visual consistency. 
Moreover, adapting generative models for this task typically relies on extensive high-quality paired audio-video editing data, which are costly to obtain. 
Building on the rich cross-modal priors of pretrained text-to-audio-video models, we propose \textbf{DAJA}, a decoupled adaptation framework for instruction-guided audio-video editing that reduces reliance on paired editing data. 
By freezing the pretrained backbone and restricting adaptation to audio tokens, DAJA learns sound editing from readily available audio-only data, requiring only a small amount of paired audio-video supervision, while exploiting existing visual capabilities and cross-modal interactions.
Beyond model adaptation, DAJA decouples multimodal edit reasoning from prompt formulation, preserving reasoning about scene dependencies while tailoring prompts to the generation model.
Experimental results demonstrate that DAJA outperforms state-of-the-art audio-video editing methods in editing fidelity, audio-visual consistency, and generation quality, highlighting decoupled adaptation as a simple and effective approach to joint editing with limited paired audio-video supervision.
Codes are available at \url{https://github.com/YujiaHu1109/DAJA}.

\end{abstract}    
\section{Introduction}
\label{sec:intro}

Recent advances in audio-video generation have demonstrated strong capabilities in synthesizing coherent visual and acoustic content~\cite{mmdiffusion,ovi,ltx2,minimaxh3}. 
Building on this progress, instruction-guided audio-video editing~\cite{aved,instructav2av,javedit,uniavc} aims to modify an existing video and its soundtrack according to an instruction. 
Beyond generating plausible content, this task requires coordinated modifications across modalities while preserving source content unrelated to the intended edit. 
For instance, replacing a sound-producing object requires modifying both the object and its associated sound, while retaining all unrelated visual elements and acoustic content.

Despite recent progress, learning coordinated audio-video editing remains challenging, as adapting generative models typically requires substantial amounts of high-quality paired audio-video editing data~\cite{instructav2av,javedit,editinsync}, which are costly to obtain. 
Furthermore, the emphasis on human speech and appearance in some existing datasets~\cite{javedit,editinsync} leaves other sound-producing entities, such as animals and objects, less extensively covered.
Consequently, these limitations motivate data-efficient adaptation strategies for learning diverse editing capabilities from limited paired audio-video data.

Pretrained text-to-audio-video models~\cite{ovi,ltx2,minimaxh3} offer a promising foundation for addressing these challenges, as they already encode rich priors relating visual content to sound.
MiniMax-H3~\cite{minimaxh3}, for example, can already modify visual content according to editing instructions. However, we find that sounds associated with the original scene often remain unchanged, even when their editing is explicitly requested.
This imbalance suggests a targeted adaptation strategy: learning audio editing from more readily available audio-only data~\cite{audit,mmedit}, while drawing on the model's existing visual capabilities and cross-modal priors to support coordinated editing.
Encouragingly, our pilot study provides preliminary support for this strategy, indicating that audio-only supervision can benefit audio modeling within a joint audio-video generator.


Building on these findings, we propose \textbf{DAJA}, a simple and parameter-efficient decoupled adaptation framework for instruction-guided joint audio-video editing. 
DAJA reuses the pretrained joint generation architecture,
freezing its backbone and localizing trainable adaptation to audio tokens while retaining shared audio-visual attention. 
This decouples audio adaptation from the visual pathway while preserving shared audio-visual interactions to exploit pretrained cross-modal priors.
Readily available audio-only editing data provide the primary supervision for learning sound transformations, complemented by only a small amount of paired audio-video data to coordinate the learned audio edits with the model's existing visual editing capabilities.

Beyond model adaptation, coherent editing requires not only accurate analysis of the intended changes and their dependencies~\cite{veggie,cotedit}, but also their formulation as structured, model-compatible generation conditions.
DAJA therefore decouples multimodal edit reasoning from prompt formulation, preserving the reasoning model's ability to analyze scene dependencies while adapting prompt formulation to the generator's requirements, including explicit audio-visual edits and content preservation constraints.

Experiments on joint audio-video editing tasks demonstrate that DAJA outperforms state-of-the-art audio-video editing methods in editing fidelity, audio-video consistency, and generation quality. 
These results show that our decoupled adaptation makes instruction-guided joint audio-video editing simple, requiring only limited paired audio-video supervision.
Our contributions are summarized as follows:

\begin{itemize}
    \item We identify the imbalance between visual and audio editability and the limited diversity in paired editing data. These findings motivate learning audio editing from readily available audio-only supervision while exploiting pretrained visual capabilities and cross-modal priors.

    \item We propose \textbf{DAJA}, a simple adaptation framework that decouples audio adaptation from the visual pathway and edit reasoning from prompt formulation, enabling coordinated audio-video editing with limited paired supervision.

    \item Experiments demonstrate that DAJA exhibits superior editing fidelity, audio-video consistency, and output quality on joint audio-video editing tasks compared with state-of-the-art audio-video editing methods.

\end{itemize}

\section{Related Work}
\label{sec:related_work}

\begin{figure*}[t]
\centering
\includegraphics[width=\textwidth]{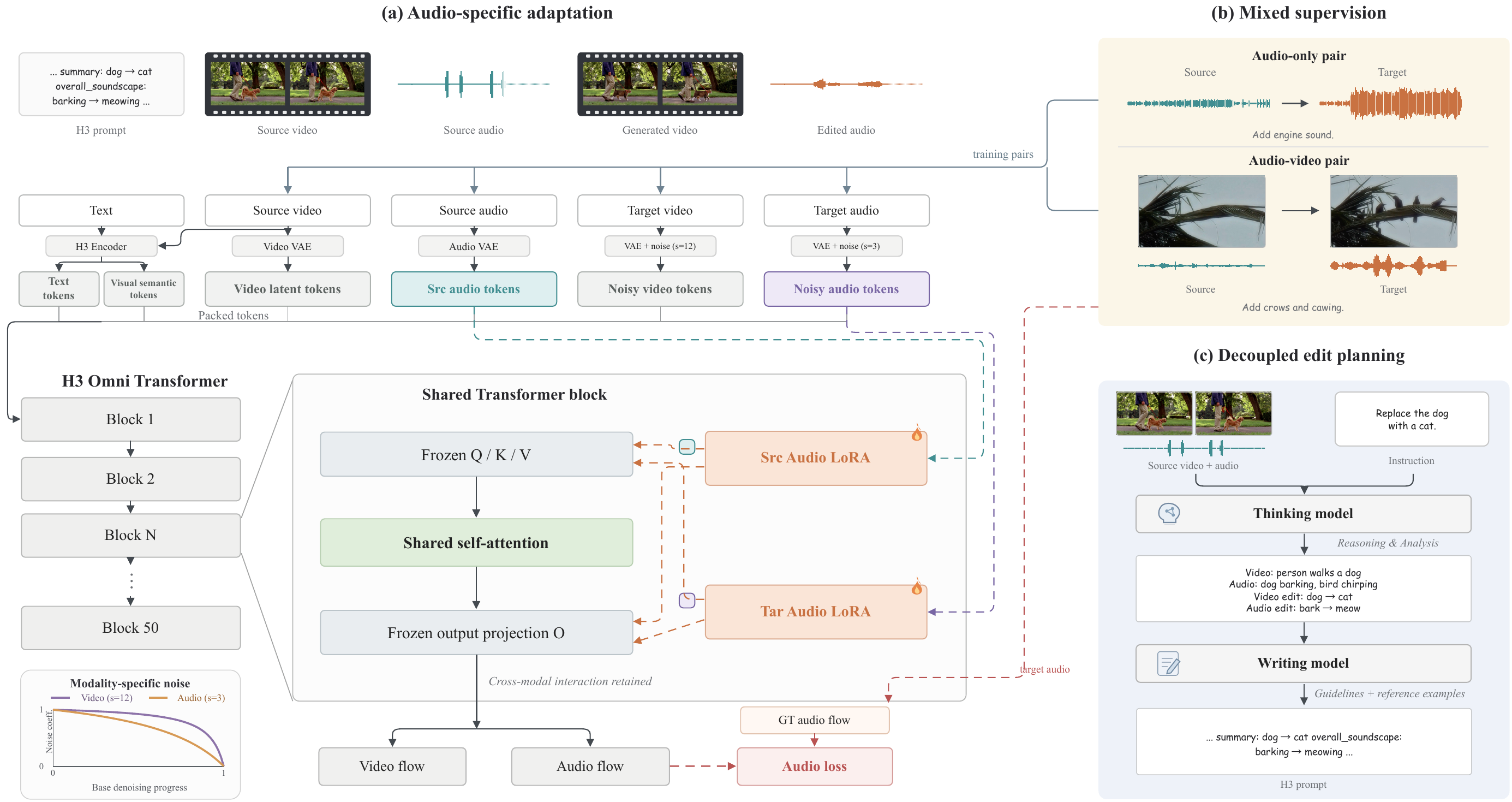}
\vspace{-3mm}
\caption{
Overview of DAJA.
(a) Audio-specific adaptation introduces separate source and target audio adapters into the frozen H3 backbone, retaining shared audio-visual attention while restricting direct adapter updates to audio tokens.
(b) Audio-only editing data, complemented by a small set of paired audio-video examples, train the adapters through target audio flow supervision.
(c) Decoupled edit planning separates multimodal reasoning from prompt formulation.
}
\label{fig:method}
\vspace{-3mm}
\end{figure*}

\subsection{Audio-Video Generation and Editing}
\label{sec:rw_av_editing}

Joint audio-video generation~\cite{mmdiffusion,ovi,ltx2,minimaxh3}
uses coupled diffusion models~\cite{mmdiffusion} and cross-modal
transformer fusion~\cite{ovi,ltx2} to synthesize synchronized
visual and acoustic content.
Single-stream architectures further support multimodal reference
conditioning through shared attention~\cite{minimaxh3}.
Joint editing approaches include one-shot image--audio
adaptation~\cite{avedit}, training-free
editing~\cite{aved,avflowedit}, object-level
manipulation~\cite{objectavedit}, and audio generation conditioned
on edited videos~\cite{coherentavedit}.
Instruction-trained models learn from curated editing
pairs~\cite{instructav2av}, including human-centric appearance
and speech modifications~\cite{javedit,editinsync}.

Audio-only editing~\cite{audit,mmedit} covers sound~\cite{audit},
music~\cite{instructme}, speech~\cite{instructspeech}, and multiple
editing operations~\cite{mmedit}, with diffusion inversion enabling
training-free edits~\cite{audioeditor}.
Cross-modal approaches reduce paired-data requirements through
audio editing combined with audio-video
reconstruction~\cite{crossedit}, or progressive unimodal training
and cross-modal alignment~\cite{uniavc}.
We focus on audio-specific adaptation using readily available
audio-only supervision and limited paired edits, while retaining
pretrained cross-modal interactions.

\subsection{Adaptation and Control for Visual Editing}
\label{sec:rw_adaptation}

Instruction-based visual editing~\cite{instructpix2pix} is
complemented by cross-attention control~\cite{prompttoprompt}
and inversion-free flow-based editing~\cite{flowedit}.
Video editing employs one-shot tuning~\cite{tuneavideo},
spectral adaptation~\cite{save}, attention
control~\cite{videop2p,fatezero}, and feature
propagation~\cite{tokenflow} for source preservation and
temporal consistency.

Low-rank adaptation~\cite{lora} provides compact parameter updates,
while control branches~\cite{controlnet} and context
adapters~\cite{vace} introduce additional conditioning.
In-context adapters support synchronized portrait
editing~\cite{synclora}, and modular adapters enable audio-visual
controls~\cite{avcontrol}.
Low-rank fine-tuning also supports joint audio-video
editing~\cite{javedit}.
Our separate source- and target-audio adapters specialize reference
conditioning and edited-audio generation, retaining shared attention
without directly adapting visual tokens.

\subsection{Multimodal Reasoning and Edit Planning}
\label{sec:rw_planning}

Multimodal language models support grounded instruction
elaboration~\cite{mgie} and reasoning-aware image
editing~\cite{smartedit}.
Other methods incorporate image and audio
references~\cite{instructany2pix}, or use thinking and reflection
to refine edits~\cite{reasonedit}.
Video editing similarly explores instruction
grounding~\cite{veggie}, multimodal guidance~\cite{instructx},
and explicit planning and reasoning~\cite{cotedit,thinkv2v}.
Audio agents provide temporal guidance for synchronized instance
editing~\cite{aviedit}.
In this paper, we separate multimodal edit reasoning from prompt formulation,
translating dependencies among entities, actions, and sounds
into explicit editing and preservation conditions.    
\section{Method}
\label{sec:method}

\subsection{Preliminaries and Motivation}
\noindent\textbf{Preliminaries.}
\label{sec:preliminary}
We build our method on MiniMax-H3 \cite{minimaxh3}, a pretrained audio-video generation model that jointly processes text, reference audio-video content, and generation targets using a shared Transformer. 
As shown in Fig.~\ref{fig:method}(a), for audio-video editing, the model takes a source video $v_s$, its audio track $a_s$, and an editing prompt $c$ as conditions, and generates an edited video and audio.

Let $\mathbf z_v^0$ and $\mathbf z_a^0$ denote the clean target video and audio latents, respectively. For each modality $m\in\{v,a\}$, the noisy latent is constructed as
\begin{equation}
\mathbf z_m^t
=
t_m\mathbf z_m^0
+
(1-t_m)\boldsymbol{\epsilon}_m,
\qquad
\boldsymbol{\epsilon}_m\sim\mathcal N(0,I),
\end{equation}
where $t_m$ controls the noise level of modality $m$ \cite{flowmatching}. Video and audio use distinct noise schedules, as illustrated in Fig.~\ref{fig:method}(a).

The text, reference, and noisy target tokens are packed into a sequence and processed by the shared Transformer:
\begin{equation}
(\hat{\mathbf u}_v,\hat{\mathbf u}_a)
=
F_{\theta_0}
\left(
\mathbf z_v^t,\mathbf z_a^t;
v_s,a_s,c,t_v,t_a
\right),
\end{equation}
where $\theta_0$ denotes the pretrained model parameters, and $\hat{\mathbf u}_v$ and $\hat{\mathbf u}_a$ are the predicted video and audio flows. 
The shared Transformer allows the two modalities to interact during generation, even when they are at different noise levels.

\begin{figure}[t]
    \centering
    \includegraphics[width=\columnwidth]{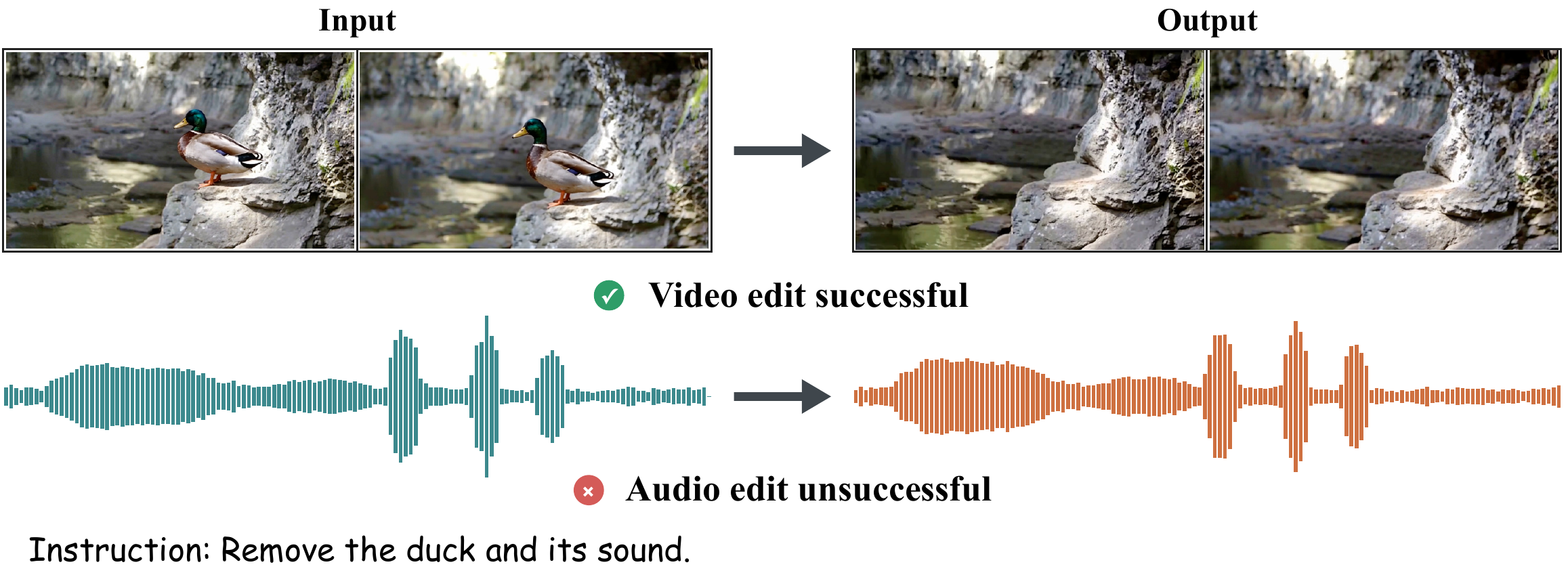}
    \caption{Imbalanced visual and audio editability in MiniMax-H3.}
    \vspace{-5mm}
    \label{fig:observation}
\end{figure}

\noindent\textbf{Observation.}
Although MiniMax-H3 has strong audio-video generation capabilities, we find that it often struggles to edit existing sounds as instructed. 
In most cases we examined, the model successfully modifies the visual content but retains the sound associated with the original object or action, even when the prompt explicitly requests its removal or replacement, as shown in Fig.~\ref{fig:observation}. 
The few successful audio edits we observed mainly involve human speech, and their results are not always consistent. 
This concentration on speech-related edits parallels the human-centric focus of existing audio-video editing datasets, such as JAVEdit~\cite{javedit}. 
As a result, accurately editing non-human content, e.g., objects, animals, and their associated sounds, remains challenging, despite its relevance to common editing scenarios. 
These observations reveal a gap between visual and audio editing capabilities.

\noindent\textbf{Motivation.}
A key challenge in learning audio-video editing is the limited availability of diverse, high-quality paired editing data~\cite{instructav2av,crossedit}.
In contrast, audio-only editing pairs can be obtained or constructed more readily across diverse sound events~\cite{audit,mmedit}.
This raises a natural question: \emph{can readily available audio-only supervision improve sound editing while retaining pretrained visual capabilities and cross-modal interactions?}

\begin{figure}[t]
    \centering
    \includegraphics[width=\columnwidth]{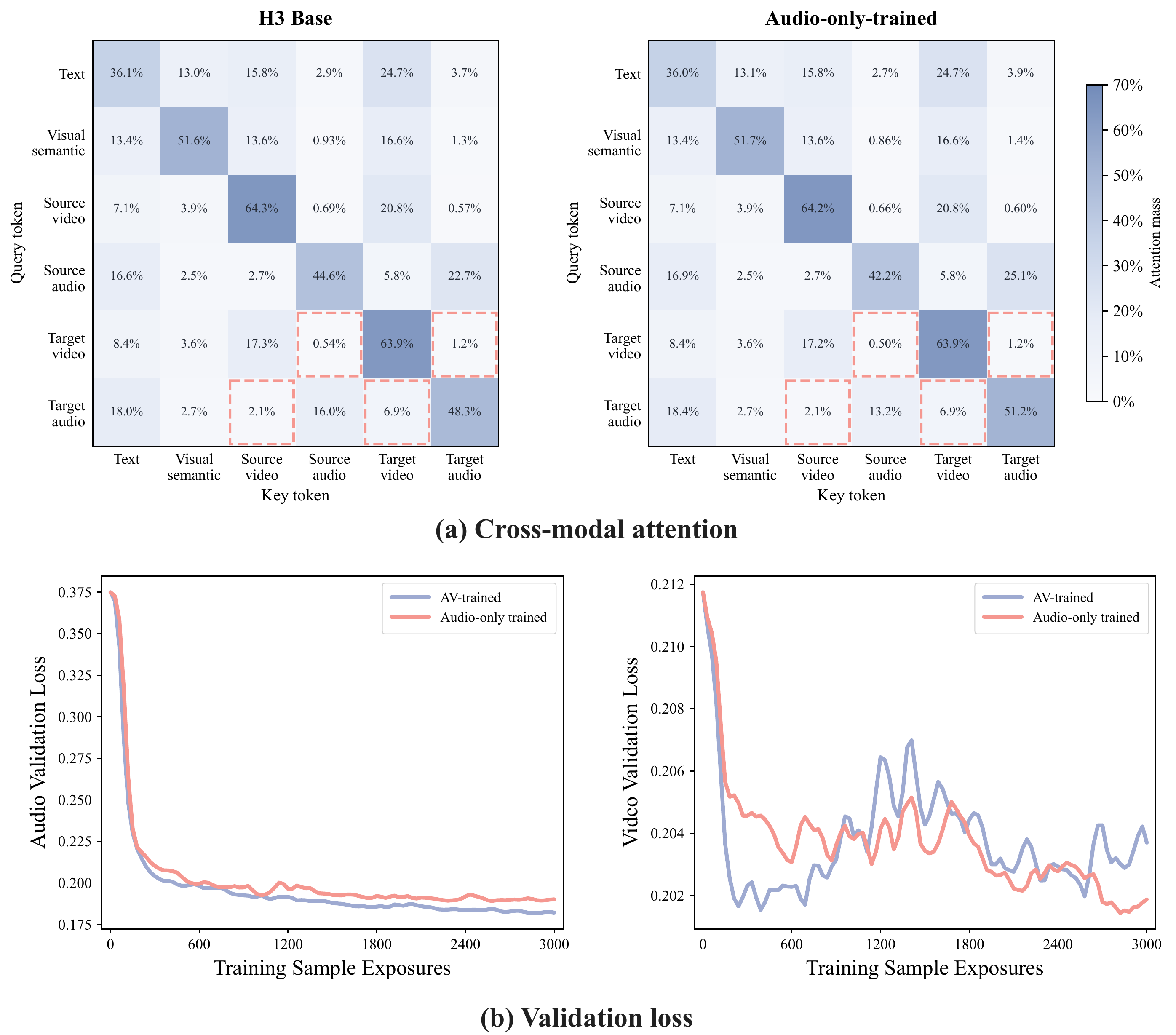}
    \caption{Empirical motivation for audio-specific adaptation.}
    \vspace{-5mm}
    \label{fig:motivation}
\end{figure}

To investigate this possibility, we examine cross-modal attention before and after audio-only adaptation. As shown in Fig.~\ref{fig:motivation}(a), target audio and video tokens
attend to tokens from the other modality in both models. Moreover, the aggregate attention patterns remain similar after adaptation. This observation suggests that audio-only adaptation can retain cross-modal attention pathways in the shared Transformer.

We further conduct a controlled pilot study using the same editing examples with and without video conditioning during training, with both models evaluated under the same audio-video conditions. As shown in Fig.~\ref{fig:motivation}(b), audio-only and audio-video training exhibit broadly similar trends in both audio and video validation loss under identical audio-video conditioning. These results provide preliminary support for learning audio-flow prediction from audio-only supervision.

Together, these findings motivate an audio-focused adaptation strategy that builds on the pretrained model's visual capabilities and cross-modal interactions to support coordinated audio-video editing.

\subsection{Decoupled Audio-Visual Adaptation}
\label{sec:audio_adaptation}
Guided by the motivation in Sec.~\ref{sec:preliminary}, we introduce decoupled audio-visual
adaptation, which retains the pretrained visual pathway and shared cross-modal interactions while learning audio editing from readily available audio-only data and limited paired audio-video supervision.

\noindent\textbf{Decoupling adaptation while retaining interaction.}
Since audio and video tokens share Transformer parameters, optimizing these parameters for audio editing can also alter visual processing.
As illustrated in Fig.~\ref{fig:method}(a), we therefore keep the pretrained backbone frozen and introduce trainable adapters exclusively for audio tokens, retaining the original shared attention.

Specifically, we introduce separate adapters for source audio reference tokens and target audio generation tokens, parameterized by $\phi_{\mathrm{src}}$ and $\phi_{\mathrm{tar}}$, respectively.
The source adapter learns how to use the original soundtrack as an editing reference, whereas the target adapter learns how to generate the edited audio consistent with the intended sound transformations. 
Only audio tokens receive adapter contributions in the attention projections; text and video tokens maintain the pretrained projections, which separates the adaptation of audio conditioning and generation from direct modification of the visual pathway.

The adapted audio representations nevertheless remain part of the original shared attention computation:
\begin{equation}
\operatorname{Attn}_i
=
\sum_{j\in\mathcal I_{\mathrm{all}}}
\alpha_{ij}\mathbf V_j,
\qquad
i\in\mathcal I_a\cup\mathcal I_v,
\label{eq:shared_attention}
\end{equation}
where $\mathcal I_a$ and $\mathcal I_v$ denote audio and video token indices, $\mathcal I_{\mathrm{all}}$ contains all token indices, and $\alpha_{ij}$ denotes the attention weight from query token $i$ to key token $j$.

Retaining this interaction is essential to exploiting the pretrained cross-modal prior. During joint generation, audio tokens can use visual context to guide sound synthesis, while video tokens can incorporate the adapted audio context to support coordinated visual and acoustic changes. 
Thus, our decoupling constrains the scope of parameter adaptation while preserving the model's capacity for context-dependent cross-modal generation. The visual pathway retains its pretrained parameters and remains responsive to the joint editing context through shared attention.

\noindent\textbf{Learning sound editing from mixed supervision.}
To reduce reliance on paired audio-video editing data, we learn sound editing through a shared objective across two complementary data sources.
As illustrated in Fig.~\ref{fig:method}(b), we combine audio-only editing pairs $\mathcal D_{\mathrm A}$ with a small set of paired audio-video edits $\mathcal D_{\mathrm{AV}}$.
Audio-only data provide the primary supervision for learning diverse sound transformations, while paired edits expose the same adapters to editing under visual conditioning.

Both data sources update the same audio adaptation parameters
$\phi=\{\phi_{\mathrm{src}},\phi_{\mathrm{tar}}\}$.
Following the flow-matching formulation in
Sec.~\ref{sec:preliminary}, we supervise the target audio flow using
\begin{equation}
\mathcal L_{\mathrm{audio}}
=
\mathbb E
\left[
\left\|
\hat{\mathbf u}_a
-
(\mathbf z_a^0-\boldsymbol{\epsilon}_a)
\right\|_2^2
\right],
\label{eq:audio_flow_loss}
\end{equation}
where the expectation is over sampled timesteps and noise.
The squared error is averaged over target audio latent elements corresponding to the original, unpadded duration; source reference tokens and temporal padding are excluded.
The adaptation objective is
\begin{equation}
\min_{\phi}\;
\mathbb E_{x\sim p_{\mathrm{mix}}}
\left[
\mathcal L_{\mathrm{audio}}(x;\theta_0,\phi)
\right],
\label{eq:adaptation_objective}
\end{equation}
where $p_{\mathrm{mix}}$ samples from both
$\mathcal D_{\mathrm A}$ and $\mathcal D_{\mathrm{AV}}$,
and $\theta_0$ remains frozen.

The shared audio flow objective concentrates adaptation on sound editing, consistent with retaining the pretrained visual pathway.
For paired examples, video tokens provide visual context through shared attention, while supervision remains on the target audio flow, as shown in Fig.~\ref{fig:method}(a).
At inference, DAJA retains the original joint denoising process to generate coordinated video and audio from the source content and editing prompt.

\subsection{Decoupled Edit Planning}
\label{sec:edit_planning}

Instruction-guided audio-video editing requires interpreting potentially underspecified requests within the source scene and expressing their implications as coherent generation conditions.
The official H3 pipeline uses Context-IR to construct an intermediate representation. Nevertheless, we find that its prompts may not fully capture dependencies among entities, actions, and sounds, or sufficiently specify the intended audio edits.
We therefore disentangle semantic edit reasoning from model-specific prompt formulation to address these complementary requirements separately, as illustrated in Fig.~\ref{fig:method}(c).

The reasoning stage derives a structured edit plan $\mathbf p$ from the source video $v_s$, audio $a_s$, and instruction $c$. We guide the reasoning model \cite{qwen3omni} to first analyze the source content and then reason about the requested edit.
The model considers how changes to an entity affect related actions and sounds, identifying the necessary edits and content to preserve across modalities.

In the formulation stage, a writing model \cite{qwen3} converts the structured plan into an H3-compatible prompt. Guided by the official writing guidelines $\mathcal G$ and representative examples $\mathcal E$, it organizes visual changes, audio operations, and preservation requirements into explicit generation conditions.
The two stages are formulated as
\begin{equation}
\mathbf p = P_{\mathrm{think}}(v_s,a_s,c),
\quad
c_{\mathrm{H3}} = P_{\mathrm{write}}(\mathbf p,\mathcal G,\mathcal E).
\end{equation}
This separation exploits the complementary strengths of reasoning and writing models, enabling scene-dependent editing intent to be expressed as coherent generation conditions.
\providecommand{\DAJATrainSteps}{\textbf{[TBD]}}
\providecommand{\DAJAInsAVECount}{\textbf{[TBD]}}
\providecommand{\dajafigureplaceholder}[2]{%
  \fbox{\begin{minipage}[c][#1][c]{0.965\linewidth}
    \centering\small\textit{#2}
  \end{minipage}}}

\section{Experiments}
\label{sec:experiments}

\begin{figure*}[t]
\centering
\includegraphics[width=\textwidth]{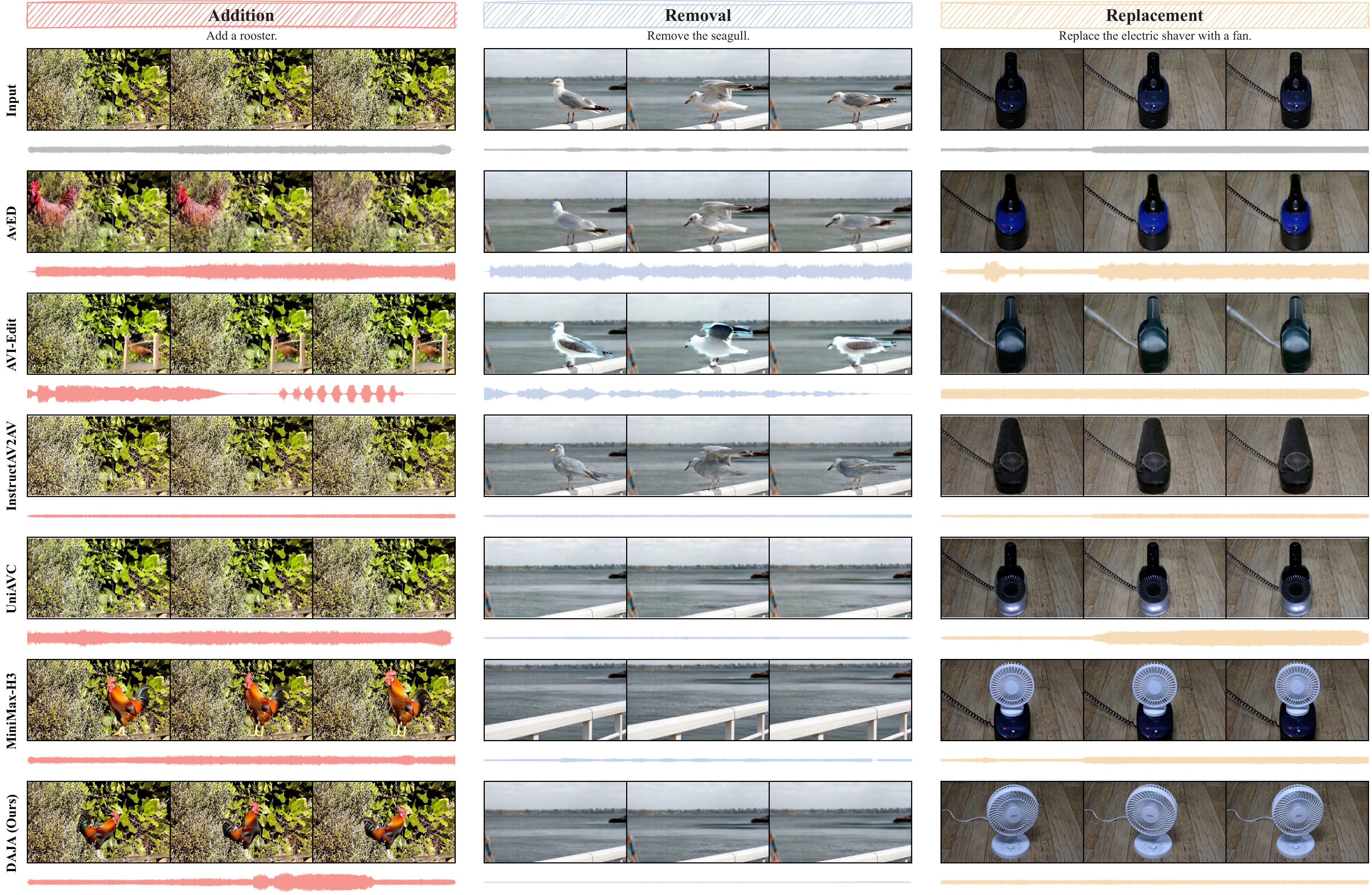}
\caption{Qualitative comparison results with existing state-of-the-art methods on joint audio-video editing tasks.}
\label{fig:main_qualitative}
\end{figure*}

\begin{table*}[t]
\centering
\caption{Quantitative comparison with state-of-the-art approaches on our test set.}
\label{tab:main_comparison}
\setlength{\tabcolsep}{4pt}
\renewcommand{\arraystretch}{1.18}
\resizebox{\textwidth}{!}{%
\begin{tabular}{l*{11}{c}}
\toprule
& \multicolumn{3}{c}{Editing fidelity and preservation}
& \multicolumn{4}{c}{Alignment and synchrony}
& \multicolumn{4}{c}{Generation quality} \\
\cmidrule(lr){2-4}
\cmidrule(lr){5-8}
\cmidrule(lr){9-12}
Method
& AV-FES $\uparrow$
& V-FES $\uparrow$
& A-FES $\uparrow$
& TV-A $\uparrow$
& TA-A $\uparrow$
& AV-A $\uparrow$
& PEAVS $\uparrow$
& IQ $\uparrow$
& MS $\uparrow$
& PQ $\uparrow$
& PAM $\uparrow$ \\
\midrule
AvED~\cite{aved}
& 0.328 & 0.410 & 0.245
& 20.77 & \textbf{26.20} & 21.08 & 3.27
& 0.497 & 0.957 & 5.913 & 0.506 \\
AVI-Edit~\cite{aviedit}
& 0.186 & 0.327 & 0.044
& 21.42 & 17.85 & 19.28 & 3.18
& 0.448 & 0.987 & 5.847 & 0.648 \\
InstructAV2AV~\cite{instructav2av}
& 0.304 & 0.296 & \underline{0.313}
& 20.76 & 17.43 & 14.42 & 3.47
& 0.514 & 0.989 & 6.163 & 0.564 \\
UniAVC~\cite{uniavc}
& 0.293 & 0.469 & 0.118
& 20.38 & 16.67 & 22.66 & 3.33
& 0.517 & 0.989 & \textbf{6.290} & 0.586 \\
\midrule
MiniMax-H3~\cite{minimaxh3}
& \underline{0.429} & \underline{0.790} & 0.067
& \underline{21.71} & \underline{25.65}
& \underline{23.12} & \underline{3.80}
& \textbf{0.545} & \underline{0.991} & 6.162
& \underline{0.653} \\
\midrule
\textbf{DAJA (Ours)}
& \textbf{0.653} & \textbf{0.840} & \textbf{0.465}
& \textbf{21.74} & \underline{25.65} & \textbf{25.41} & \textbf{3.88}
& \underline{0.540} & \textbf{0.992} & \underline{6.283}
& \textbf{0.660} \\
\bottomrule
\end{tabular}}
\end{table*}

\subsection{Experimental Setup}
\label{sec:exp_setup}

\noindent\textbf{Training settings.}
We initialize our model from MiniMax-H3 Ref2VA~\cite{minimaxh3} and freeze its pretrained parameters. Separate source- and target-audio LoRA adapters~\cite{lora} are introduced into the query, key, value, and output projections of all 50 Omni-Transformer blocks, with
rank $r=64$ and scaling parameter $\alpha=64$.
We enable each LoRA adapter only for its corresponding audio-token type, with LoRA disabled for text and video tokens.
Our training uses target audio flow-matching supervision, with audio-only and audio-video examples sampled at an $8{:}2$ ratio. Videos are processed at a resolution of $576\times320$ and 24 FPS, with 124 frames per clip, and temporally aligned audio is resampled to 32 kHz.
We train for 10 epochs using AdamW with a learning rate of $5\times10^{-5}$ and cosine
learning-rate decay. 

\noindent\textbf{Dataset setup.}
Our training set comprises approximately 7K audio-only and 600 audio-video editing pairs covering addition, removal, and replacement. 
The audio-only data combine pairs from MMEdit-TestSet~\cite{mmedit}
with edits synthesized using resources from ESC-50~\cite{esc50} and MMEdit-TestSet.
For audio-video pairs, we curate source videos from VGGSound~\cite{vggsound} and JAVEdit~\cite{javedit}, formulate editing instructions using GPT-6 Astra~\cite{gpt6astra}, and generate target videos with Seedance 2.5~\cite{seedance25}, followed by quality checks.
We further supplement these constructed pairs with existing examples from JAVEdit and InsAVE-80K~\cite{instructav2av}.

For evaluation, we construct a held-out test set of 180 instruction-guided audio-video editing tasks, separate from the training set, with 60 tasks each for addition, removal, and replacement. 
Each task comprises a source audio-video clip and an editing instruction formulated using GPT-6 Astra~\cite{gpt6astra}.
The original source clips span approximately 4--15 seconds, predominantly at a resolution of $1280\times720$. 

\noindent\textbf{Evaluation settings.}
We compare DAJA with five representative audio-video editing methods: AvED~\cite{aved}, AVI-Edit~\cite{aviedit}, InstructAV2AV~\cite{instructav2av}, UniAVC~\cite{uniavc}, and MiniMax-H3~\cite{minimaxh3}. 
For MiniMax-H3, we use the official API to generate the corresponding H3 prompt. 
For DAJA, we adopt two-stage planning as illustrated in Sec.\ref{sec:edit_planning} with Qwen3-Omni-Thinking \cite{qwen3omni} first analyzing the source audio-video and editing instruction, and Qwen3-8B \cite{qwen3} translating the structured analysis into an H3 prompt.

We evaluate editing fidelity and content preservation using V-FES, A-FES, and AV-FES~\cite{crossedit} for video, audio, and their aggregate performance, respectively.
Instruction alignment is measured using text-video alignment (TV-A) based on VideoCLIP-XL~\cite{videoclipxl} and text-audio alignment (TA-A) based on CLAP~\cite{clap}.
We assess cross-modal consistency through audio-video alignment (AV-A) using ImageBind~\cite{imagebind} and perceptual synchrony using PEAVS~\cite{peavs}.
We additionally report Imaging Quality and Motion Smoothness from VBench~\cite{vbench} to assess video quality, and Production Quality (PQ) from Audiobox Aesthetics~\cite{audiobox_aesthetics} and PAM~\cite{pam} to assess audio quality. 
More experimental details are provided in the appendix.

\subsection{Comparison with Existing Methods}
\label{sec:exp_main}

\begin{figure}[t]
    \centering
    \includegraphics[width=\columnwidth]{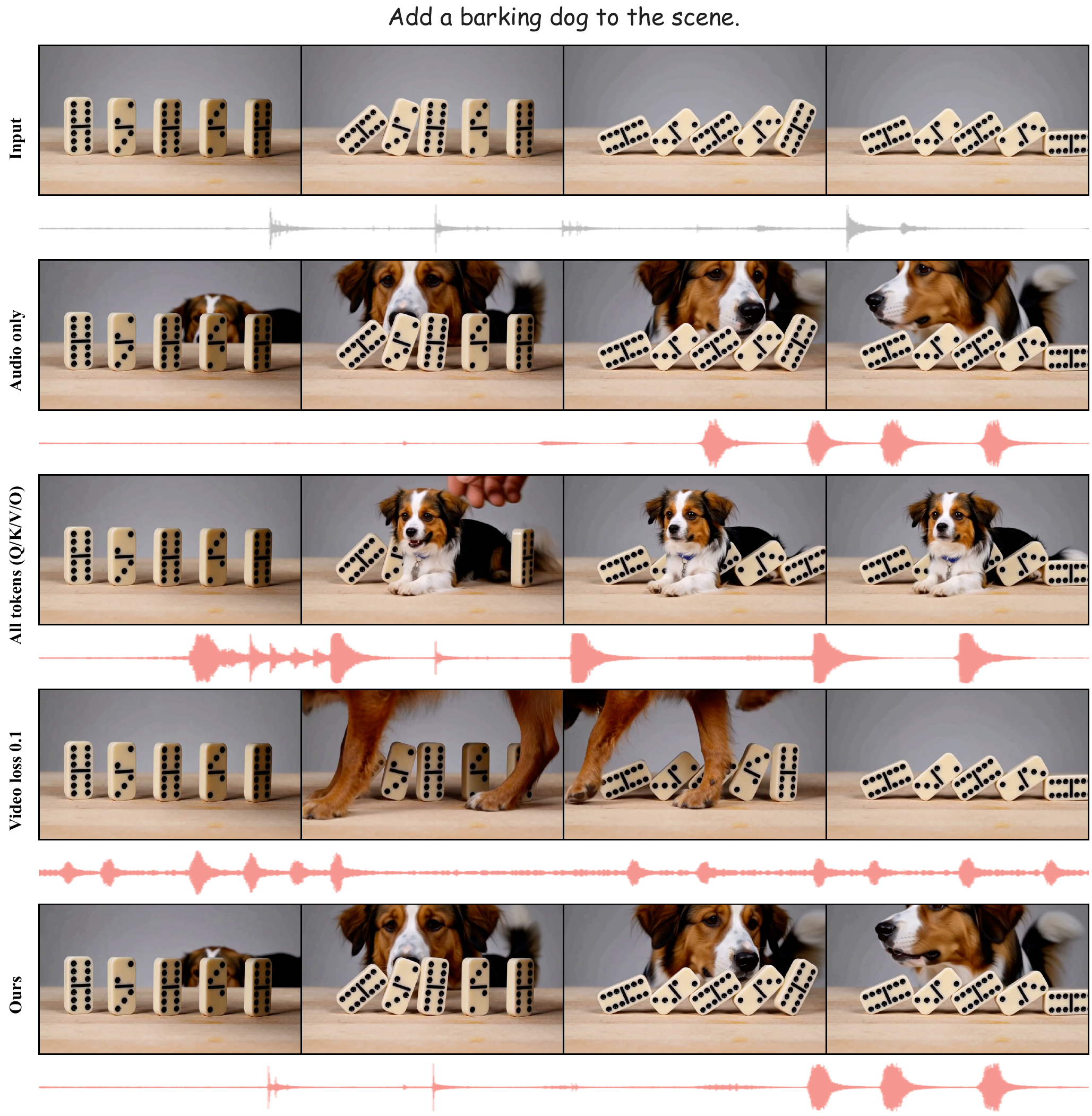}
    \vspace{-3mm}
    \caption{Qualitative ablation results of audio-visual adaptation.}
    \vspace{-3mm}
    \label{fig:adaptation_ablation}
\end{figure}

\noindent\textbf{Quantitative Comparisons.} Tab.~\ref{tab:main_comparison} exhibits the quantitative comparison results of our DAJA and other approaches on our test set. The results demonstrate that DAJA achieves the best performance overall in editing fidelity, audio-visual alignment, and generation quality. 
Compared with the backbone model MiniMax-H3, DAJA substantially improves A-FES from 0.067 to 0.465 while also increasing V-FES from 0.790 to 0.840, demonstrating improved audio editing without compromising visual editing performance.

\noindent\textbf{Qualitative Comparisons.}
Fig.~\ref{fig:main_qualitative} presents qualitative comparison results of DAJA against others for editing tasks including subject addition, removal and replacement. As shown in Fig.~\ref{fig:main_qualitative}, unlike other competing methods that often leave audio edits incomplete or introduce unintended changes, DAJA performs the requested visual and audio edits while preserving unrelated content and maintaining high generation quality. More comparison results are provided in the appendix.

\subsection{Ablation Studies}
\label{sec:exp_ablation}

\begin{figure}[t]
    \centering
    \includegraphics[width=\columnwidth]{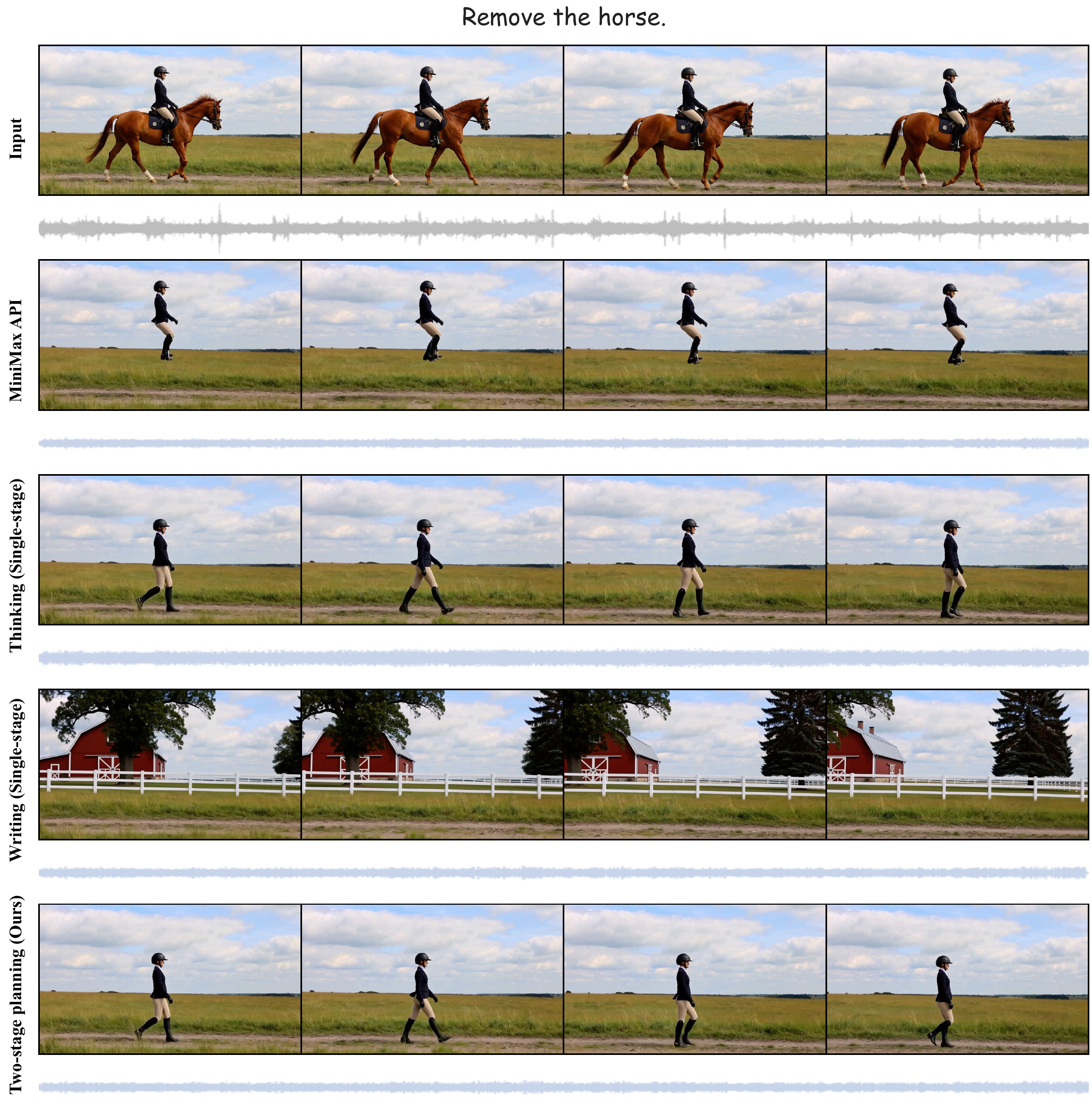}
    \vspace{-3mm}
    \caption{Qualitative ablation results of edit planning strategies.}
    \label{fig:planning_ablation}
    \vspace{-3mm}
\end{figure}

We investigate the effects of training data mixture, LoRA configuration, video supervision, and edit planning on our test set. The quantitative ablation results are summarized in Tab.~\ref{tab:ablation}. More ablation details are provided in the appendix.

\begin{table*}[t]
\centering
\caption{Quantitative ablation results on data mixture, LoRA placement, video loss and edit planning.}
\label{tab:ablation}
\setlength{\tabcolsep}{3pt}
\renewcommand{\arraystretch}{1.2}

\resizebox{\textwidth}{!}{%
\begin{tabular}{ll*{11}{c}}
\toprule
& Setting
& AV-FES $\uparrow$
& V-FES $\uparrow$
& A-FES $\uparrow$
& TV-A $\uparrow$
& TA-A $\uparrow$
& AV-A $\uparrow$
& PEAVS $\uparrow$
& IQ $\uparrow$
& MS $\uparrow$
& PQ $\uparrow$
& PAM $\uparrow$ \\
\midrule

\multirow{4}{*}{\textbf{Data mixture}}
& AV-only
& 0.602 & 0.780 & 0.423
& \underline{21.72} & \underline{25.71} & 25.26
& \textbf{3.91} & \underline{0.538} & \textbf{0.994}
& 6.194 & 0.655 \\

& Audio-only
& 0.622 & 0.776 & \textbf{0.468}
& 21.46 & \textbf{25.85} & 23.88
& 3.49 & 0.533 & 0.989
& \textbf{6.288} & 0.658 \\

& Audio:AV $= 9{:}1$
& \underline{0.646} & \underline{0.828} & 0.463
& 21.68 & 25.62 & \underline{25.38}
& 3.86 & \underline{0.538} & \underline{0.992}
& \underline{6.283} & \textbf{0.662} \\

& \cellcolor{tableblue}\textbf{Audio:AV $= 8{:}2$ (Ours)}
& \cellcolor{tableblue}\textbf{0.653}
& \cellcolor{tableblue}\textbf{0.840}
& \cellcolor{tableblue}\underline{0.465}
& \cellcolor{tableblue}\textbf{21.74}
& \cellcolor{tableblue}25.65
& \cellcolor{tableblue}\textbf{25.41}
& \cellcolor{tableblue}\underline{3.88}
& \cellcolor{tableblue}\textbf{0.540}
& \cellcolor{tableblue}\underline{0.992}
& \cellcolor{tableblue}\underline{6.283}
& \cellcolor{tableblue}\underline{0.660} \\
\midrule

\multirow{4}{*}{\textbf{LoRA placement}}
& All tokens, Q/K/V/O
& 0.536 & 0.657 & 0.414
& 20.88 & 25.33 & 24.87
& 3.55 & 0.529 & 0.987
& 6.204 & 0.641 \\

& Tgt audio, Q/K/V/O
& \underline{0.642} & \underline{0.825} & \underline{0.458}
& 21.63 & 25.59 & \underline{25.38}
& 3.72 & \textbf{0.542} & \textbf{0.992}
& \textbf{6.285} & \underline{0.653} \\

& Src+Tgt audio, Q/O
& 0.627 & 0.810 & 0.443
& \underline{21.70} & \textbf{25.68} & 25.33
& \underline{3.86} & 0.535 & \underline{0.991}
& 6.189 & 0.642 \\

& \cellcolor{tableblue}\textbf{Src+Tgt audio, Q/K/V/O (Ours)}
& \cellcolor{tableblue}\textbf{0.653}
& \cellcolor{tableblue}\textbf{0.840}
& \cellcolor{tableblue}\textbf{0.465}
& \cellcolor{tableblue}\textbf{21.74}
& \cellcolor{tableblue}\underline{25.65}
& \cellcolor{tableblue}\textbf{25.41}
& \cellcolor{tableblue}\textbf{3.88}
& \cellcolor{tableblue}\underline{0.540}
& \cellcolor{tableblue}\textbf{0.992}
& \cellcolor{tableblue}\underline{6.283}
& \cellcolor{tableblue}\textbf{0.660} \\
\midrule

\multirow{2}{*}{\textbf{Video loss}}
& $\lambda_v = 0.1$
& 0.595 & 0.726 & 0.463
& 21.39 & 25.62 & 24.03
& 3.57 & 0.526 & 0.988
& 6.279 & 0.658 \\

& \cellcolor{tableblue}$\bm{\lambda_v = 0}$ \textbf{(Ours)}
& \cellcolor{tableblue}\textbf{0.653}
& \cellcolor{tableblue}\textbf{0.840}
& \cellcolor{tableblue}\textbf{0.465}
& \cellcolor{tableblue}\textbf{21.74}
& \cellcolor{tableblue}\textbf{25.65}
& \cellcolor{tableblue}\textbf{25.41}
& \cellcolor{tableblue}\textbf{3.88}
& \cellcolor{tableblue}\textbf{0.540}
& \cellcolor{tableblue}\textbf{0.992}
& \cellcolor{tableblue}\textbf{6.283}
& \cellcolor{tableblue}\textbf{0.660} \\
\midrule

\multirow{4}{*}{\textbf{Edit planning}}
& MiniMax API
& 0.505 & \underline{0.817} & 0.192
& \underline{21.68} & 25.32 & 23.89
& \underline{3.81} & \textbf{0.544} & \underline{0.991}
& 6.189 & 0.653 \\

& Thinking (single-stage)
& \underline{0.621} & 0.799 & \underline{0.442}
& 21.33 & \underline{25.53} & \underline{25.33}
& 3.79 & 0.525 & 0.990
& \underline{6.251} & 0.648 \\

& Writing (single-stage)
& 0.271 & 0.457 & 0.085
& 19.62 & 20.98 & 25.10
& 3.76 & 0.503 & 0.985
& 6.202 & \underline{0.655} \\

& \cellcolor{tableblue}\textbf{Two-stage planning (Ours)}
& \cellcolor{tableblue}\textbf{0.653}
& \cellcolor{tableblue}\textbf{0.840}
& \cellcolor{tableblue}\textbf{0.465}
& \cellcolor{tableblue}\textbf{21.74}
& \cellcolor{tableblue}\textbf{25.65}
& \cellcolor{tableblue}\textbf{25.41}
& \cellcolor{tableblue}\textbf{3.88}
& \cellcolor{tableblue}\underline{0.540}
& \cellcolor{tableblue}\textbf{0.992}
& \cellcolor{tableblue}\textbf{6.283}
& \cellcolor{tableblue}\textbf{0.660} \\
\bottomrule
\vspace{-6mm}
\end{tabular}}
\end{table*}

\noindent\textbf{Effect of the training data mixture.} We evaluate four training data configurations: AV-only, audio-only, and audio-to-AV sampling ratios of $9{:}1$ and $8{:}2$.
As shown in Tab.~\ref{tab:ablation}, AV-only training yields lower A-FES than mixed-data training, potentially due to the limited scale of paired audio-video editing data.
Audio-only training compromises audio-visual consistency, as illustrated in Fig.~\ref{fig:adaptation_ablation}, where a dog bark is generated without corresponding mouth movement.
Combining the two data sources provides diverse audio supervision alongside visual conditioning, improving overall performance.
The $8{:}2$ mixture outperforms $9{:}1$ on most metrics and achieves the highest AV-FES among the four configurations.

\noindent\textbf{Effect of LoRA configurations.} We compare four LoRA configurations: Q/K/V/O adaptation for all tokens, Q/K/V/O for target audio tokens, Q/O for source and target audio tokens, and our source-and-target audio Q/K/V/O configuration.
Applying LoRA to all tokens yields lower scores across the reported metrics and may introduce unintended visual changes, as shown in Fig.~\ref{fig:adaptation_ablation}. Adapting only target-audio tokens lowers performance on most metrics, suggesting that source-audio conditioning also benefits from adaptation.
Similarly, restricting both audio token groups to Q/O projections reduces performance, potentially because omitting K/V adaptation limits how audio context is accessed and propagated through shared attention.
Together, these findings support adapting both source and target audio tokens through Q/K/V/O projections.

\noindent\textbf{Effect of video supervision.} We examine whether additional visual supervision benefits joint editing by adding a target video flow loss with weight $\lambda_v=0.1$ to the default audio-flow objective. We can observe in Tab.~\ref{tab:ablation} that adding a video loss decreases all reported metrics and leads to visible degradation in the generated frames, as shown in Fig.~\ref{fig:adaptation_ablation}. This supports focusing the adaptation objective on target audio flow without additional video supervision.

\begin{figure}[t]
    \centering
    \includegraphics[width=\columnwidth]{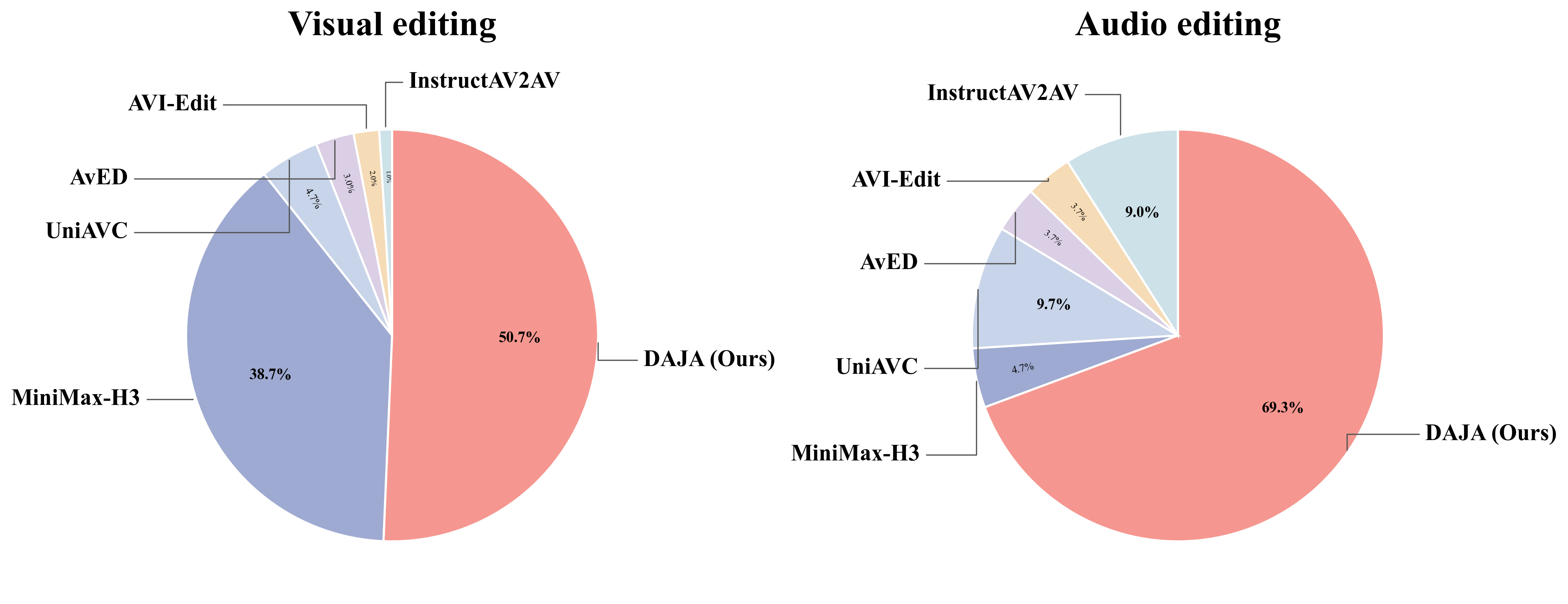}
    \vspace{-3mm}
    \caption{User study results for visual and audio editing.}
    \vspace{-3mm}
    \label{fig:user_study}
\end{figure}

\noindent\textbf{Effect of edit planning.} We compare different H3 prompt generation approaches, including MiniMax API, single-stage thinking, single-stage writing, and our two-stage planner. Representative qualitative results are shown in  Fig.~\ref{fig:planning_ablation}. 
MiniMax API achieves slightly higher image quality but performs worse on the other metrics. Its prompts may overlook scene dependencies or omit explicit audio edits. 
Single-stage thinking also underperforms, potentially reflecting the difficulty of jointly handling scene analysis and generator-specific prompt formulation within a single model. 
Meanwhile, single-stage writing produces well-formatted prompts but cannot directly perceive the source video or audio, limiting its ability to ground the requested edits in the source scene.
Our two-stage planner analyzes scene dependencies before writing the prompt. As shown in Fig.~\ref{fig:planning_ablation}, with our two-stage prompts, the generator removes the horse and its associated sound while adjusting the person's pose and motion to walking on the ground.

\subsection{User Study}
\label{sec:exp_user_study}

We conduct a user study with 20 participants and 15 examples covering addition, removal, and replacement. Given the source audio-video clip, editing instruction, and anonymized results in randomized order, participants are asked to separately select the best result for visual and auditory editing.
The results are summarized in Fig.~\ref{fig:user_study}. DAJA receives the largest share of votes on the 15 evaluated examples: 50.7\% for visual editing and 69.3\% for audio editing. Its particularly large lead in auditory editing highlights the effectiveness of our approach.
\section{Conclusions}
\label{sec:conclu}
We presented DAJA, a decoupled adaptation framework for instruction-guided joint audio-video editing.
By restricting adaptation to audio tokens within a frozen backbone, DAJA learns sound editing from readily available audio-only data while exploiting pretrained visual capabilities and cross-modal priors.
We further decouple edit reasoning from prompt formulation, preserving reasoning about scene dependencies while tailoring prompts to the generator.
Experiments and human evaluation demonstrate improved editing fidelity, audio-video consistency, and generation quality with limited paired audio-video supervision.

\section*{LLM Usage Statement}
LLMs and other generative AI tools, including GPT-6, were used to assist with editing and refining author-produced manuscript text, related-work discussions, code and implementation, experimental design and analysis, figures, and the presentation of research ideas and conclusions.

\small
\bibliographystyle{ieeenat_fullname}
\bibliography{main}

\appendix
\clearpage
\setcounter{page}{1}
\maketitlesupplementary


\providecommand{\verify}[1]{\textcolor{red}{\textbf{[To verify:} #1\textbf{]}}}

\section{Implementation Details}
\label{sec:supp_implementation}

The model and training setup are summarized in Tab.~\ref{tab:supp_config}.

\begin{table*}[t]
\centering
\small

\caption{Implementation settings.}
\label{tab:supp_config}
\begin{tabularx}{\textwidth}{@{}lX@{}}
\toprule
Setting & Configuration \\
\midrule
Backbone
& MiniMax-H3 Ref2VA~\cite{minimaxh3};
pretrained parameters frozen \\

Adapters
& Source- and target-audio LoRA~\cite{lora},
Q/K/V/O in all 50 blocks; $r=\alpha=64$, dropout $0$ \\

Trainable parameters
& 321,126,400 total; 160,563,200 per audio role \\

Objective / data sampling
& Target audio flow loss; audio-only : audio-video $=8{:}2$ \\

Optimizer
& AdamW, learning rate $5\times10^{-5}$, cosine decay \\

Input media
& Video $576\times320$ at 24 FPS; audio 32 kHz \\

Training budget
& 10 epochs \\

Inference
& 50 denoising steps with MiniMaxH3Scheduler,
seed $42$, and no classifier-free guidance \\
\bottomrule
\end{tabularx}

\vspace{0.8em}

\caption{Input composition and supervision for audio-only
and audio-video training examples.}
\label{tab:supp_tokens}
\begin{tabularx}{\textwidth}{@{}lXX@{}}
\toprule
Input & Audio-only pair & Audio-video pair \\
\midrule
Text
& H3-formatted audio-editing prompt
& H3-formatted audio-video editing prompt \\

Visual semantic tokens
& None 
& Source-video semantic features \\

Video latent tokens
& None
& Source video and noised target video \\

Audio latent tokens
& Source audio and noised target audio
& Source audio and noised target audio \\

Supervision
& Target audio flow
& Target audio flow; no video-flow loss \\
\bottomrule
\end{tabularx}

\vspace{0.8em}

\caption{Configuration of the two-stage planner. Both models
use BF16, SDPA, and deterministic decoding.}
\label{tab:supp_planner}
\begin{tabularx}{\textwidth}{@{}lXX@{}}
\toprule
Setting & Thinking & Writing \\
\midrule
Model
& Qwen3-Omni-30B-A3B-Thinking~\cite{qwen3omni}
& Qwen3-8B Base~\cite{qwen3} \\

Inputs
& Video frames, synchronized audio, original instruction
& Instruction, edit type, reference metadata, analysis,
guide, examples \\

Media / decoding
& 2 FPS, maximum 262,144 pixels/frame; audio read from video
& Text only; \texttt{enable\_thinking=False} \\

Generation
& \texttt{do\_sample=False}
& \texttt{do\_sample=False} \\
\bottomrule
\end{tabularx}

\end{table*}

\paragraph{Adapter routing.}
Each packed token is assigned a role. In every Transformer block, source
and target audio tokens activate their respective adapters in the Q, K, V,
and output projections. The adapter parameters are independent across
roles, projections, and layers, while text, visual, and padding tokens use
only the frozen projections. The output-projection gate follows the role
of the query token. This routing preserves cross-modal attention. The LoRA
A matrices use Kaiming initialization, and the B matrices are initialized
to zero.

\paragraph{Training inputs.}
We use H3-compatible prompts generated by our two-stage planning
pipeline as text conditions for training. Tab.~\ref{tab:supp_tokens}
summarizes the input composition. Audio-only examples contain no video
latents, whereas audio-video examples contain source and noised target
video latents. Both types contain source audio and noised target audio.
The loss is averaged over non-padding target-audio elements.

\subsection{Two-Stage Edit Planning}
\label{sec:supp_edit_planning}

The two-stage planner first analyzes the source audio-video clip and the
original instruction using a Thinking model~\cite{qwen3omni}. It then
passes the structured analysis to a text-only Writing model~\cite{qwen3}.
Tab.~\ref{tab:supp_planner} summarizes the configuration, and
Sec.~\ref{sec:supp_prompts} presents the complete prompts.

The Thinking model distinguishes observed source content from the
requested edit and considers dependencies among entities, actions, and
sounds. It returns five fields: \texttt{video\_content},
\texttt{audio\_content}, \texttt{video\_changes},
\texttt{audio\_changes}, and \texttt{uncertainty}. Audio is read from the
video using \texttt{use\_audio\_in\_video=True}, and the implementation
checks that audio features are present. The 4,096-token budget covers the
entire generation, including reasoning; the requested final JSON contains
approximately 150 English words or fewer. 

The Writing model converts the analysis into an H3 prompt with six sections:
\texttt{subject\_definitions}, \texttt{summary},
\texttt{retention\_analysis}, \texttt{detailed\_description},
\texttt{overall\_soundscape}, and \texttt{non\_diegetic\_music}. Its
system guide contains three additional examples. The
writer does not directly observe the source media.

\subsection{Preliminary Experiment Protocol}
\label{sec:supp_pilot}

The pilot compares training with and without video conditioning on the
same 180 examples, using 20 separate examples for validation. Both models
start from H3 with newly initialized audio adapters and optimize only the
audio flow objective. Each model is trained for 3K updates. Validation is
performed every ten updates under audio-video conditioning at
$u\in\{0.25,0.5,0.75\}$, yielding 60 observations per checkpoint. 

\begin{figure}[h]
    \centering

    \includegraphics[width=\columnwidth]{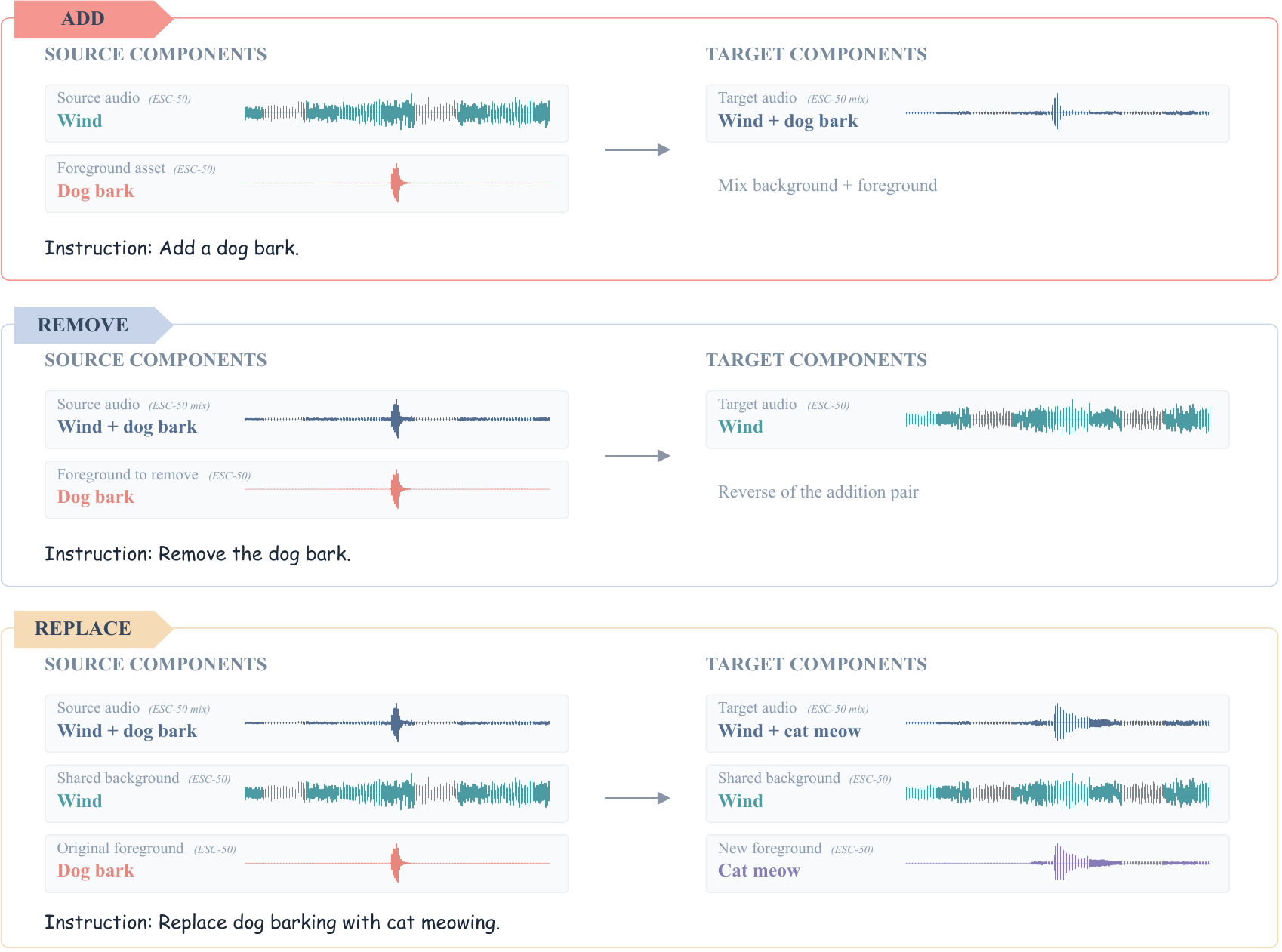}
    \caption{Construction of audio-only editing data.}
    \label{fig:audiocons}

    \vspace{2mm} 

    \includegraphics[width=\columnwidth]{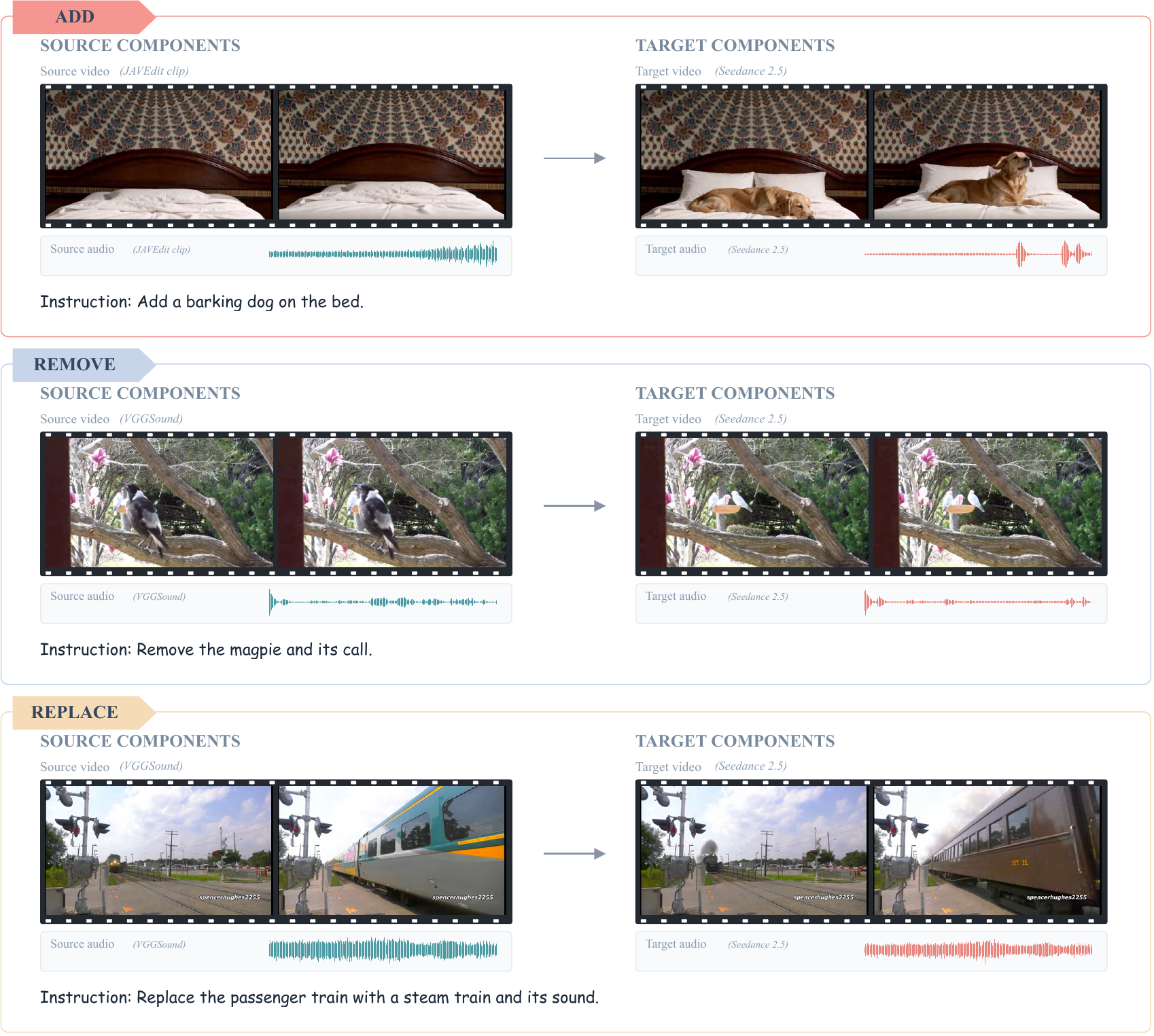}
    \caption{Construction of audio-video editing data.}
    \label{fig:avcons}
     \vspace{-5mm}

\end{figure}

For the attention comparison, we evaluate 20 validation examples
at three noise levels before and after audio-only adaptation.
Attention weights are aggregated by token group and averaged over
sampled queries, attention heads, all 50 layers, and examples.
Each entry represents the total attention mass assigned to a
key-token group; the displayed entries need not sum to 100\%
because other tokens are omitted.

\section{Dataset Construction and Splits}
\label{sec:supp_data}

Training uses approximately 7K audio-only and 600 audio-video editing
pairs. The audio-only pairs comprise examples from
MMEdit-TestSet~\cite{mmedit} and synthesized pairs using
ESC-50~\cite{esc50} and MMEdit~\cite{mmedit}.
Fig.~\ref{fig:audiocons} illustrates the synthesis workflow.

The audio-video pairs combine existing examples from JAVEdit~\cite{javedit}
and InsAVE-80K~\cite{instructav2av} with generated edits of
VGGSound~\cite{vggsound} and JAVEdit clips. For the generated edits,
GPT-6 Astra~\cite{gpt6astra} formulates the instructions, and
Seedance 2.5~\cite{seedance25} generates the target videos, which are
then checked for quality. Fig.~\ref{fig:avcons} illustrates the
construction workflow.

The data cover a diverse range of sound-producing entities and
events. In particular, approximately 2.5K audio-only and 230
audio-video pairs involve animals or insects, while approximately
1.5K and 50 pairs, respectively, involve vehicles or engines.
Other examples cover natural sounds, everyday objects, machinery,
human voices, and musical instruments.

\section{Evaluation Protocols}
\label{sec:supp_evaluation}

\begin{table*}[t]
\centering
\caption{Comparison of the official MiniMax API and our
two-stage planner with the original and adapted H3 models.}
\label{tab:supp_planner_model}
\setlength{\tabcolsep}{3pt}
\renewcommand{\arraystretch}{1.2}

\resizebox{\textwidth}{!}{%
\begin{tabular}{l*{11}{c}}
\toprule
Setting
& AV-FES $\uparrow$
& V-FES $\uparrow$
& A-FES $\uparrow$
& TV-A $\uparrow$
& TA-A $\uparrow$
& AV-A $\uparrow$
& PEAVS $\uparrow$
& IQ $\uparrow$
& MS $\uparrow$
& PQ $\uparrow$
& PAM $\uparrow$ \\
\midrule

API + H3-Base
& 0.429
& 0.790
& 0.067
& 21.71
& \underline{25.65}
& 23.12
& 3.80
& \textbf{0.545}
& \underline{0.991}
& 6.162
& 0.653 \\

API + H3-adapted
& \underline{0.505}
& \underline{0.817}
& \underline{0.192}
& 21.68
& 25.32
& \underline{23.89}
& 3.81
& \underline{0.544}
& 0.991
& \underline{6.189}
& 0.653 \\

Two-stage + H3-Base
& 0.434
& 0.796
& 0.071
& \textbf{21.76}
& \textbf{25.67}
& 23.22
& \underline{3.83}
& \underline{0.544}
& 0.990
& 6.167
& \underline{0.655} \\

\cellcolor{tableblue}\textbf{Two-stage + H3-adapted (DAJA)}
& \cellcolor{tableblue}\textbf{0.653}
& \cellcolor{tableblue}\textbf{0.840}
& \cellcolor{tableblue}\textbf{0.465}
& \cellcolor{tableblue}\underline{21.74}
& \cellcolor{tableblue}\underline{25.65}
& \cellcolor{tableblue}\textbf{25.41}
& \cellcolor{tableblue}\textbf{3.88}
& \cellcolor{tableblue}0.540
& \cellcolor{tableblue}\textbf{0.992}
& \cellcolor{tableblue}\textbf{6.283}
& \cellcolor{tableblue}\textbf{0.660} \\

\bottomrule
\end{tabular}%
}
\end{table*}

\subsection{Evaluation Metrics}
\label{sec:supp_metrics}

\paragraph{Faithful Edit Score.}
Following CrossEdit~\cite{crossedit}, we evaluate instruction following
and source consistency for each modality $m \in \{v,a\}$ on a 1--5
scale. We use Qwen3-Omni-30B-A3B-Instruct~\cite{qwen3omni} as the
evaluator and score video and audio separately. Video is sampled at 2 FPS, and audio is
converted to 16 kHz mono. Source ratings are computed once
and shared across all methods.

Let $I_m^{\mathrm{src}}$ and $I_m^{\mathrm{out}}$ denote the
instruction-following ratings of the source and output,
respectively, and let $C_m$ denote the output's consistency
with the source. The normalized instruction gain and
consistency are
\begin{equation}
g_m =
\begin{cases}
\operatorname{clip}_{[0,1]}\!\left(
\frac{I_m^{\mathrm{out}}-I_m^{\mathrm{src}}}
{5-I_m^{\mathrm{src}}}
\right), & I_m^{\mathrm{src}} < 5,\\
\mathbf{1}[C_m \geq 4], & I_m^{\mathrm{src}} = 5,
\end{cases}
\qquad
c_m = \frac{C_m-1}{4}.
\end{equation}
The per-modality score is the harmonic mean of instruction
gain and consistency:
\begin{equation}
\mathrm{FES}_m =
\begin{cases}
\frac{2g_mc_m}{g_m+c_m}, & g_m+c_m > 0,\\
0, & g_m+c_m = 0.
\end{cases}
\end{equation}
We report $\mathrm{V\text{-}FES}=\mathrm{FES}_v$,
$\mathrm{A\text{-}FES}=\mathrm{FES}_a$, and
$\mathrm{AV\text{-}FES}=
(\mathrm{FES}_v+\mathrm{FES}_a)/2$. Scores are computed
per example and then averaged over the evaluation set.

\paragraph{Alignment and synchronization.}
We compute text--video alignment (TV-A), text--audio
alignment (TA-A), and audio--video alignment (AV-A) using
VideoCLIP-XL~\cite{videoclipxl},
CLAP~\cite{clap}, and ImageBind-Huge~\cite{imagebind},
respectively. Both TV-A and TA-A use the original editing
instruction, which is fixed across methods for each example.
For TV-A, we uniformly sample eight frames and resize them to
$224\times224$. For TA-A, we extract 48 kHz mono audio and use
the LAION-CLAP 630k AudioSet fusion checkpoint. For AV-A, we
compare the generated video with its corresponding generated
audio using three 2-second clips and 16 kHz mono audio.
We assess perceptual audio-visual synchronization using
PEAVS~\cite{peavs}.

\paragraph{Generation quality.}
We measure imaging quality (IQ) with MUSIQ-SPAQ and motion
smoothness (MS) with AMT-S, following the implementations in
VBench~\cite{vbench}. IQ is computed over all decoded frames
and averaged per video. We measure audio production quality
(PQ) with Audiobox Aesthetics~\cite{audiobox_aesthetics} and
audio quality with PAM~\cite{pam}. For PAM, audio is converted
to 44.1 kHz mono and processed in 7-second segments.

All metrics are first computed per example and then averaged
across examples. 

\subsection{User Study}
\label{sec:supp_user_study}

\begin{figure}[t]
    \centering
    \includegraphics[width=\columnwidth]{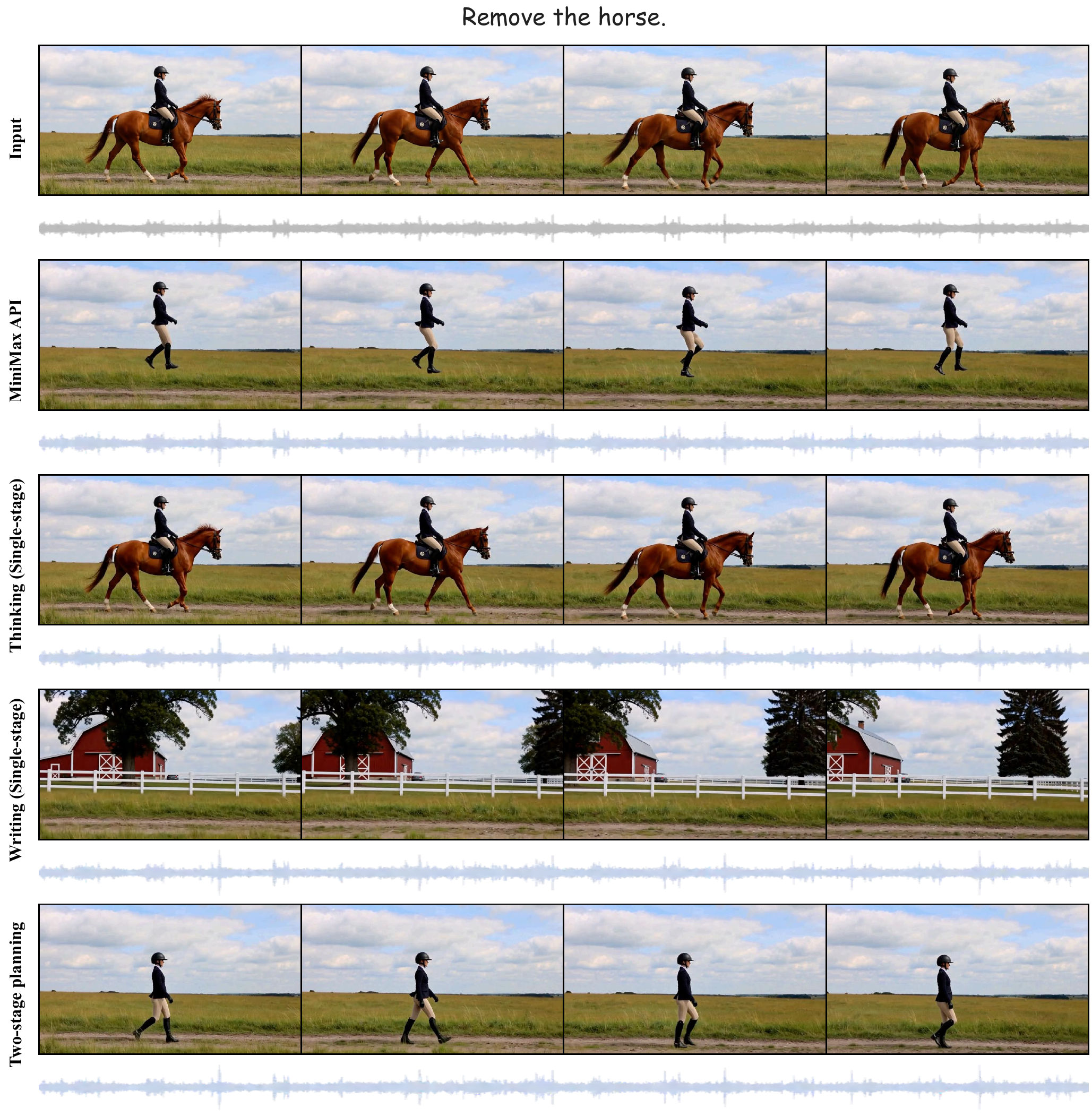}
    \vspace{-3mm}
    \caption{Qualitative ablation results of edit planning strategies on MiniMax-H3.}
    \label{fig:supp_h3_base_planning}
    \vspace{-3mm}
\end{figure}


\begin{table*}[t]
\centering
\caption{Quantitative comparison by editing type on our test set.
Each task contains 60 examples.}
\label{tab:supp_comparison_by_type}
\setlength{\tabcolsep}{4pt}
\renewcommand{\arraystretch}{1.18}

\resizebox{\textwidth}{!}{%
\begin{tabular}{c|l|ccc|cccc|cccc}
\toprule
&
& \multicolumn{3}{c|}{Editing fidelity and preservation}
& \multicolumn{4}{c|}{Alignment and synchrony}
& \multicolumn{4}{c}{Generation quality} \\

Task
& Method
& AV-FES $\uparrow$
& V-FES $\uparrow$
& A-FES $\uparrow$
& TV-A $\uparrow$
& TA-A $\uparrow$
& AV-A $\uparrow$
& PEAVS $\uparrow$
& IQ $\uparrow$
& MS $\uparrow$
& PQ $\uparrow$
& PAM $\uparrow$ \\
\midrule

\multirow{6}{*}{Addition}
& AvED~\cite{aved}
& 0.162 & 0.224 & \underline{0.101}
& 18.54 & \textbf{20.61} & 15.71 & 2.61
& 0.542 & 0.964 & 6.000 & 0.503 \\

& AVI-Edit~\cite{aviedit}
& 0.124 & 0.247 & 0.000
& 19.09 & 9.65 & 15.82 & 2.75
& 0.459 & 0.989 & 6.187 & 0.574 \\

& InstructAV2AV~\cite{instructav2av}
& 0.086 & 0.133 & 0.040
& 17.64 & 16.91 & 12.81 & 3.15
& 0.545 & 0.993 & 6.301 & 0.501 \\

& UniAVC~\cite{uniavc}
& 0.187 & 0.299 & 0.075
& 18.77 & 9.26 & \underline{22.51} & 3.25
& \textbf{0.594} & 0.993 & \textbf{6.563} & 0.588 \\

& MiniMax-H3~\cite{minimaxh3}
& \underline{0.403} & \underline{0.731} & 0.075
& \underline{20.89} & 16.00 & 20.49 & \underline{3.43}
& \underline{0.554} & \underline{0.993} & \underline{6.385} & \underline{0.640} \\

& \textbf{DAJA (Ours)}
& \cellcolor{tableblue}\textbf{0.489} & \cellcolor{tableblue}\textbf{0.746} & \cellcolor{tableblue}\textbf{0.231}
& \cellcolor{tableblue}\textbf{20.95} & \cellcolor{tableblue}\underline{18.69} & \cellcolor{tableblue}\textbf{25.41} & \cellcolor{tableblue}\textbf{3.82}
& \cellcolor{tableblue}0.552 & \cellcolor{tableblue}\textbf{0.993} & \cellcolor{tableblue}6.131 & \cellcolor{tableblue}\textbf{0.686} \\
\midrule

\multirow{6}{*}{Removal}
& AvED~\cite{aved}
& 0.281 & 0.377 & 0.186
& 21.20 & 23.46 & \underline{21.69} & 3.61
& 0.452 & 0.956 & 5.757 & 0.390 \\

& AVI-Edit~\cite{aviedit}
& 0.226 & 0.368 & 0.083
& \textbf{22.61} & 20.24 & 20.71 & 3.51
& 0.436 & 0.986 & 5.695 & \textbf{0.639} \\

& InstructAV2AV~\cite{instructav2av}
& 0.394 & 0.321 & \underline{0.466}
& \underline{22.16} & 12.21 & 12.88 & 3.63
& 0.490 & 0.988 & 5.935 & 0.514 \\

& UniAVC~\cite{uniavc}
& 0.404 & 0.646 & 0.163
& 20.50 & 17.72 & 20.37 & 3.34
& 0.467 & 0.986 & \textbf{6.135} & 0.465 \\

& MiniMax-H3~\cite{minimaxh3}
& \underline{0.443} & \underline{0.815} & 0.071
& 21.60 & \textbf{30.10} & \textbf{23.06} & \textbf{4.03}
& \textbf{0.505} & \underline{0.990} & 5.989 & \underline{0.576} \\

& \textbf{DAJA (Ours)}
& \cellcolor{tableblue}\textbf{0.734} & \cellcolor{tableblue}\textbf{0.903} & \cellcolor{tableblue}\textbf{0.565}
& \cellcolor{tableblue}21.56 & \cellcolor{tableblue}\underline{25.20} & \cellcolor{tableblue}21.69 & \cellcolor{tableblue}\underline{3.93}
& \cellcolor{tableblue}\underline{0.498} & \cellcolor{tableblue}\textbf{0.991} & \cellcolor{tableblue}\underline{6.091} & \cellcolor{tableblue}0.539 \\
\midrule

\multirow{6}{*}{Replacement}
& AvED~\cite{aved}
& \underline{0.539} & 0.630 & \underline{0.448}
& 22.56 & \textbf{34.52} & \underline{25.84} & 3.58
& 0.498 & 0.951 & 5.982 & 0.625 \\

& AVI-Edit~\cite{aviedit}
& 0.208 & 0.366 & 0.050
& 22.56 & 23.67 & 21.31 & 3.28
& 0.448 & 0.986 & 5.659 & 0.731 \\

& InstructAV2AV~\cite{instructav2av}
& 0.433 & 0.433 & 0.432
& 22.49 & 23.16 & 17.59 & 3.63
& 0.507 & 0.988 & \underline{6.253} & 0.677 \\

& UniAVC~\cite{uniavc}
& 0.289 & 0.463 & 0.115
& 21.87 & 23.04 & 25.09 & 3.40
& 0.491 & 0.988 & 6.172 & 0.705 \\

& MiniMax-H3~\cite{minimaxh3}
& 0.441 & \underline{0.825} & 0.056
& \underline{22.64} & 30.85 & 25.81 & \textbf{3.93}
& \textbf{0.576} & \underline{0.991} & 6.113 & \underline{0.741} \\

& \textbf{DAJA (Ours)}
& \cellcolor{tableblue}\textbf{0.735} & \cellcolor{tableblue}\textbf{0.870} & \cellcolor{tableblue}\textbf{0.600}
& \cellcolor{tableblue}\textbf{22.72} & \cellcolor{tableblue}\underline{33.06} & \cellcolor{tableblue}\textbf{29.14} & \cellcolor{tableblue}\underline{3.89}
& \cellcolor{tableblue}\underline{0.571} & \cellcolor{tableblue}\textbf{0.991} & \cellcolor{tableblue}\textbf{6.628} & \cellcolor{tableblue}\textbf{0.754} \\

\bottomrule
\end{tabular}%
}
\end{table*}

We conduct a user study with 20 annotators and 15 audio-video
editing cases. For each case, annotators separately select
their preferred results among six methods for visual and
audio editing, yielding 300 votes per dimension. Evaluating
the two modalities separately allows us to examine whether
improvements in visual editing are accompanied by
corresponding improvements in audio editing.

DAJA receives 152 visual-editing votes, corresponding to a
preference rate of 50.7\%. The 95\% confidence interval ranges
from 36.3\% to 65.0\%. For audio editing, DAJA receives
208 votes, corresponding to a preference rate of 69.3\%.
The 95\% confidence interval ranges from 59.0\% to 79.3\%.
DAJA obtains the highest observed preference rate in both
dimensions. Its larger share of audio-editing votes further
highlights the benefit of audio-specific adaptation.

To account for variation across annotators and cases, we
estimate confidence intervals using a two-way bootstrap.
Annotators and cases are independently resampled with
replacement, and preference rates are recomputed over
100,000 resamples. The 2.5th and 97.5th percentiles define
the reported 95\% confidence intervals.

\section{Additional Experiments and Analyses}
\label{sec:supp_additional_experiments}

\subsection{Effect of Adaptation under the Same Edit Planner}
\label{sec:supp_base_planner}

To disentangle the effects of edit planning and model
adaptation, we evaluate different prompt-generation
strategies with the original MiniMax-H3 and compare the
original and adapted models under the same two-stage planner.
This comparison allows us to examine the contribution of
each component to the final editing performance.

\paragraph{Qualitative comparison.}
Fig.~\ref{fig:supp_h3_base_planning} shows the results of
the original H3 model under different prompt settings for the
same editing example. Although the two-stage planner provides
explicit instructions for both visual and audio changes, the
original model may still retain the sound associated with the
removed object. This observation suggests that a detailed
prompt can specify the intended audio operation, but the
generator must also be able to execute that operation on the
source soundtrack.

\paragraph{Quantitative comparison.}
Tab.~\ref{tab:supp_planner_model} compares the original and
adapted H3 models under the API and two-stage planning
settings. On the original H3 model, replacing the API prompt
with the two-stage planner increases AV-FES from 0.429 to
0.434 and A-FES from 0.067 to 0.071. These changes indicate
that improved planning alone provides limited gains in audio
editing for the original model.

Under the same two-stage planner, adapting H3 increases
AV-FES from 0.434 to 0.653 and A-FES from 0.071 to 0.465,
while V-FES increases from 0.796 to 0.840. In addition, the
adapted model benefits from the two-stage planner: its
AV-FES increases from 0.505 with API prompts to 0.653 with
two-stage planning. Together, these results support the
complementary roles of the two components: the planner
formulates the intended visual and audio changes, while
audio-specific adaptation improves the generator's ability
to execute the audio edits.

\subsection{Results by Editing Type}
\label{sec:supp_edit_type}

To examine the effect of DAJA across different editing
operations, Tab.~\ref{tab:supp_comparison_by_type} reports
results separately for addition, removal, and replacement,
with 60 test examples in each subset. DAJA achieves the
highest AV-FES and A-FES across all three editing types,
demonstrating its ability to perform a range of coordinated
audio-video edits.

For addition, DAJA achieves an A-FES of 0.231, compared with
0.101 for the strongest baseline. This setting requires the
model to introduce a new sound-producing subject and its
associated audio while retaining the original scene. For
removal, DAJA achieves an A-FES of 0.565, compared with
0.466 for the strongest baseline. The task requires both the
target subject and its associated sound to be removed while
unrelated content is preserved. For replacement, DAJA
achieves an A-FES of 0.600, compared with 0.448 for the
strongest baseline. This operation requires the model to
replace both the visual subject and its corresponding sound
while maintaining their cross-modal correspondence.

These results show improvements across all three operations.
The lower A-FES for addition also indicates that introducing
new sound events remains more challenging than modifying
existing ones.

\section{More Qualitative Results}
\label{sec:supp_qualitative}

\begin{figure*}[p]
    \centering
    \includegraphics[width=\textwidth]{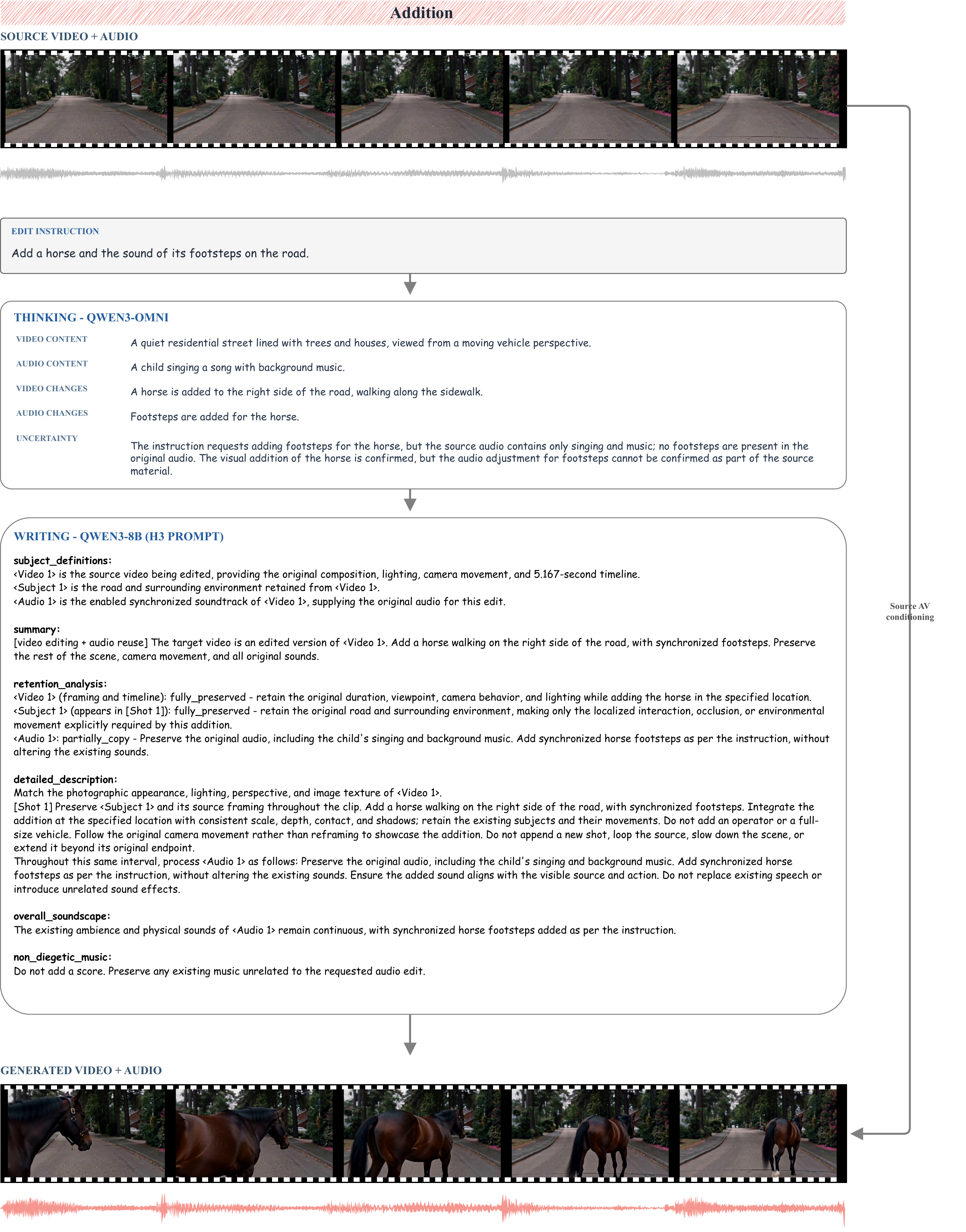}
    \caption{Additional qualitative results of DAJA: adding a horse
    and its footsteps to the road.}
    \label{fig:supp_qualitative_1}
\end{figure*}

\begin{figure*}[p]
    \centering
    \includegraphics[width=\textwidth]{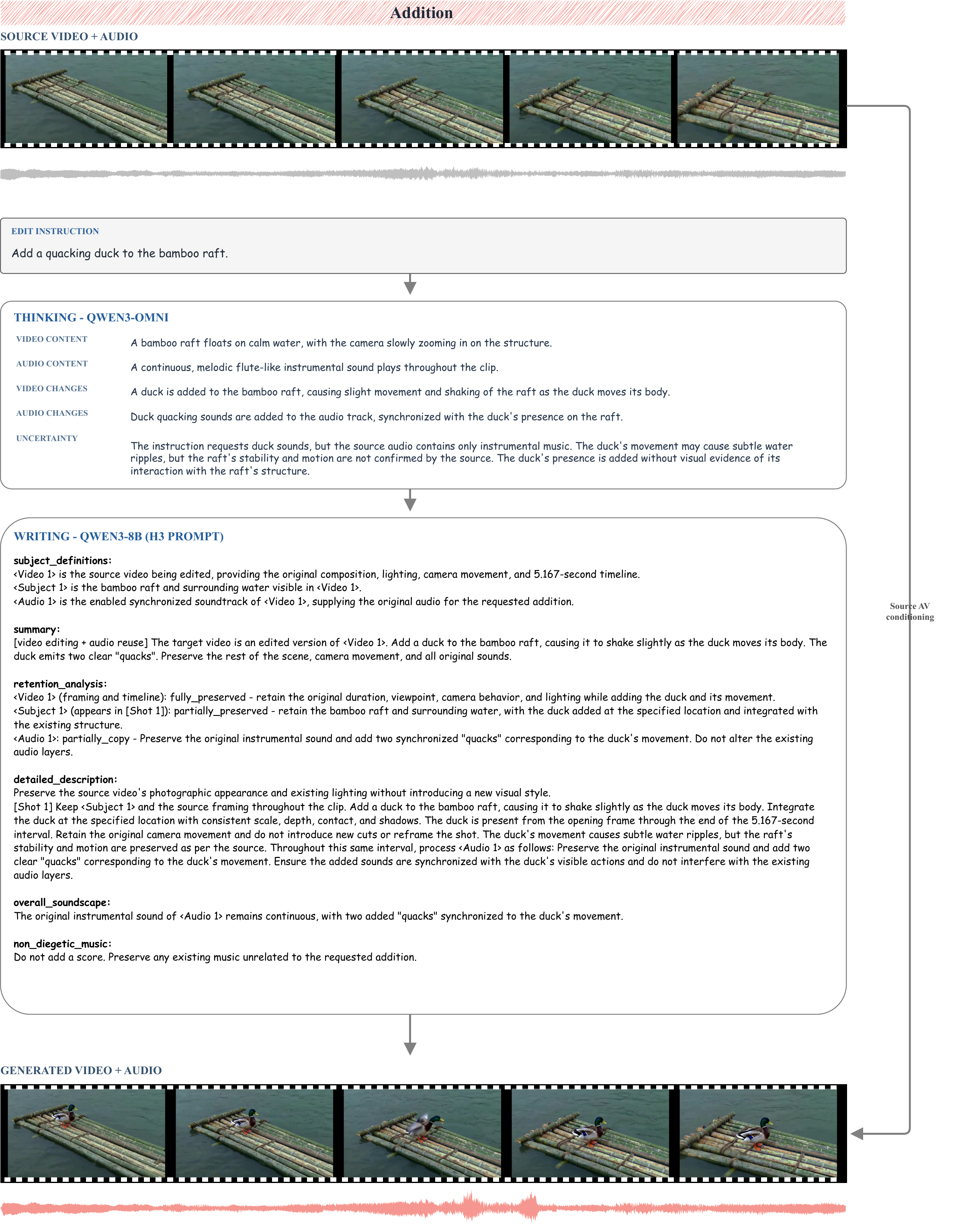}
    \caption{Additional qualitative results of DAJA: adding a duck
    and its quacking to the bamboo raft.}
    \label{fig:supp_qualitative_2}
\end{figure*}

\begin{figure*}[p]
    \centering
    \includegraphics[width=\textwidth]{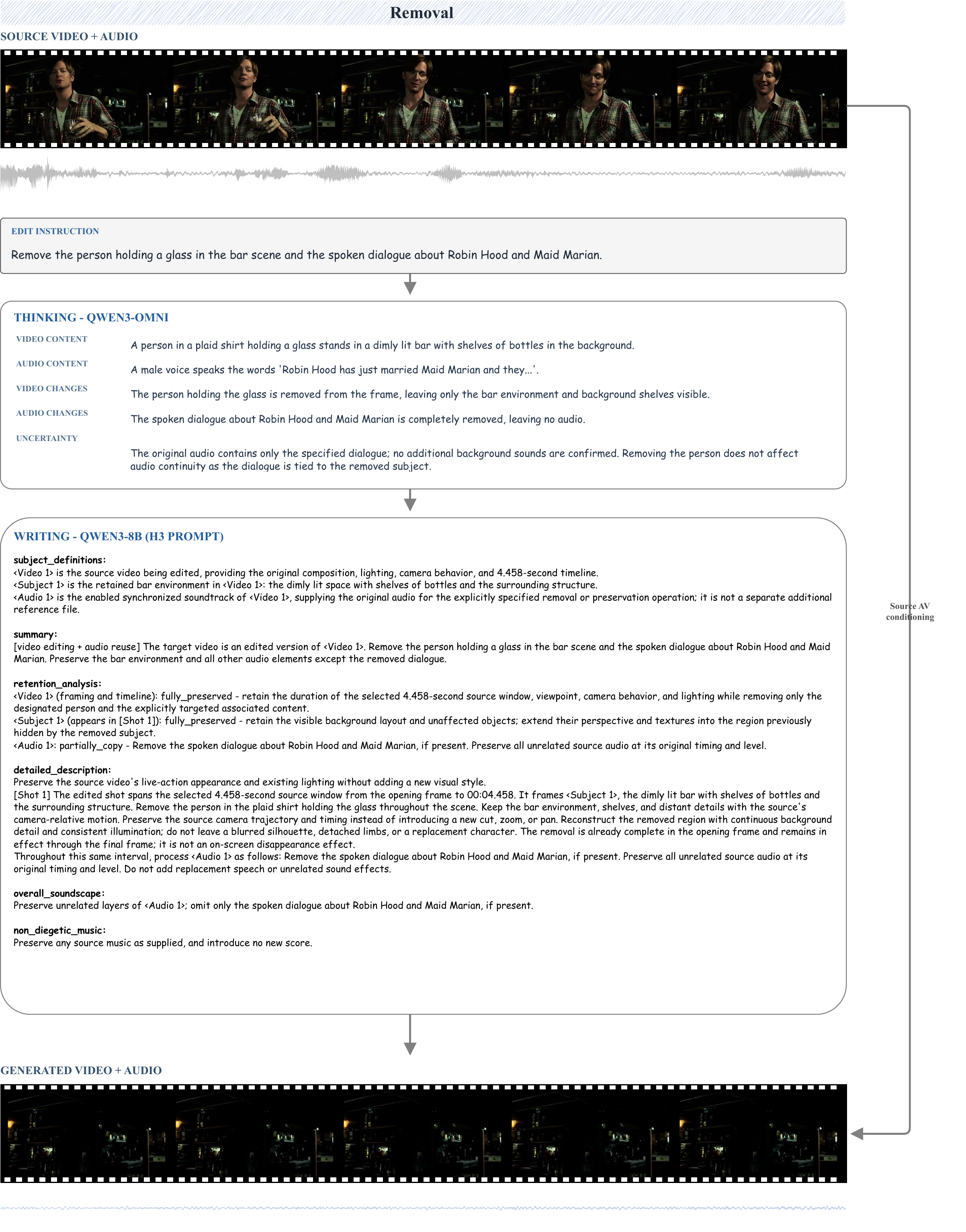}
    \caption{Additional qualitative results of DAJA: removing a
    person and his speech from the scene.}
    \label{fig:supp_qualitative_3}
\end{figure*}

\begin{figure*}[p]
    \centering
    \includegraphics[width=\textwidth]{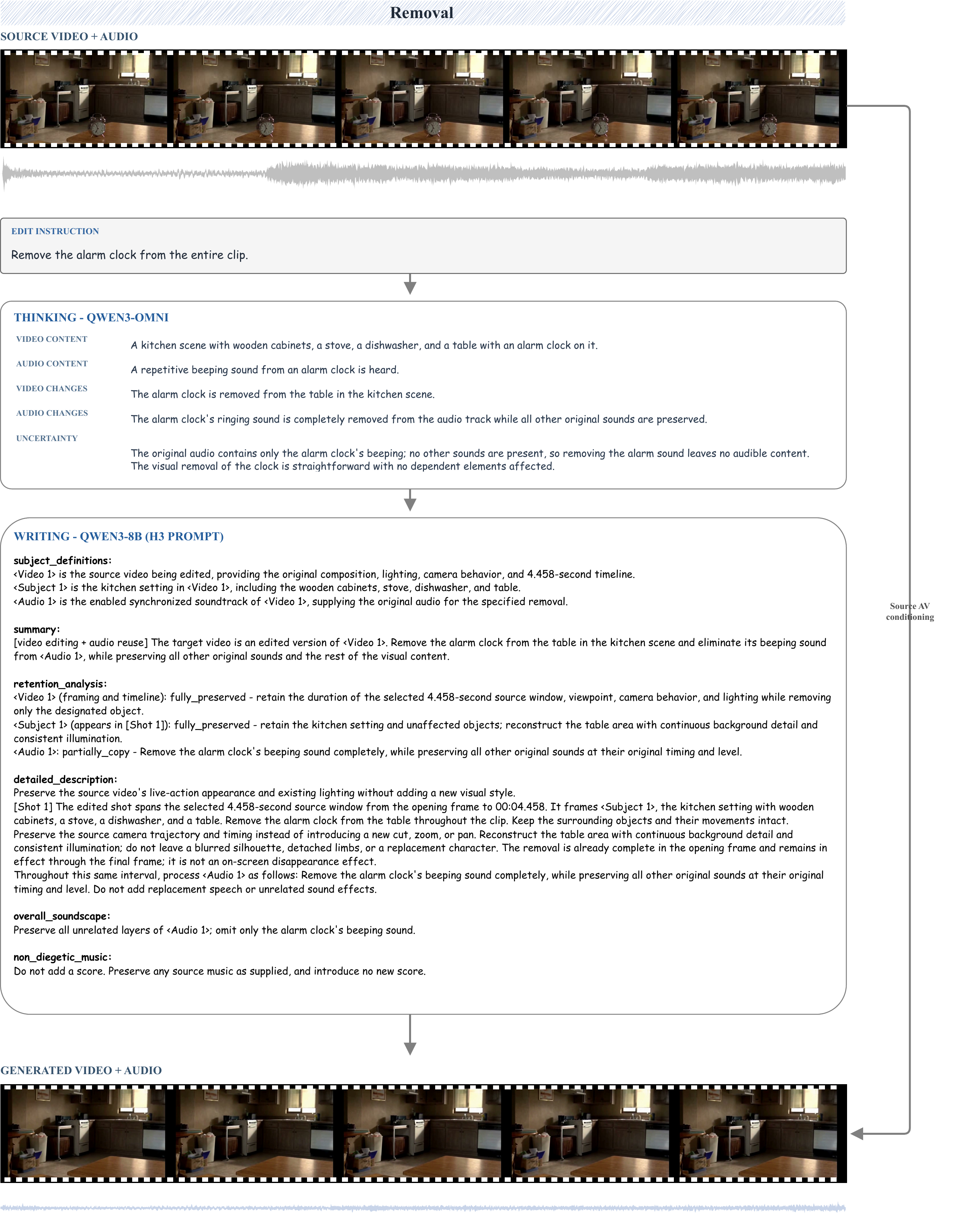}
    \caption{Additional qualitative results of DAJA: removing an
    alarm clock and its ringing from the scene.}
    \label{fig:supp_qualitative_4}
\end{figure*}

\begin{figure*}[p]
    \centering
    \includegraphics[width=\textwidth]{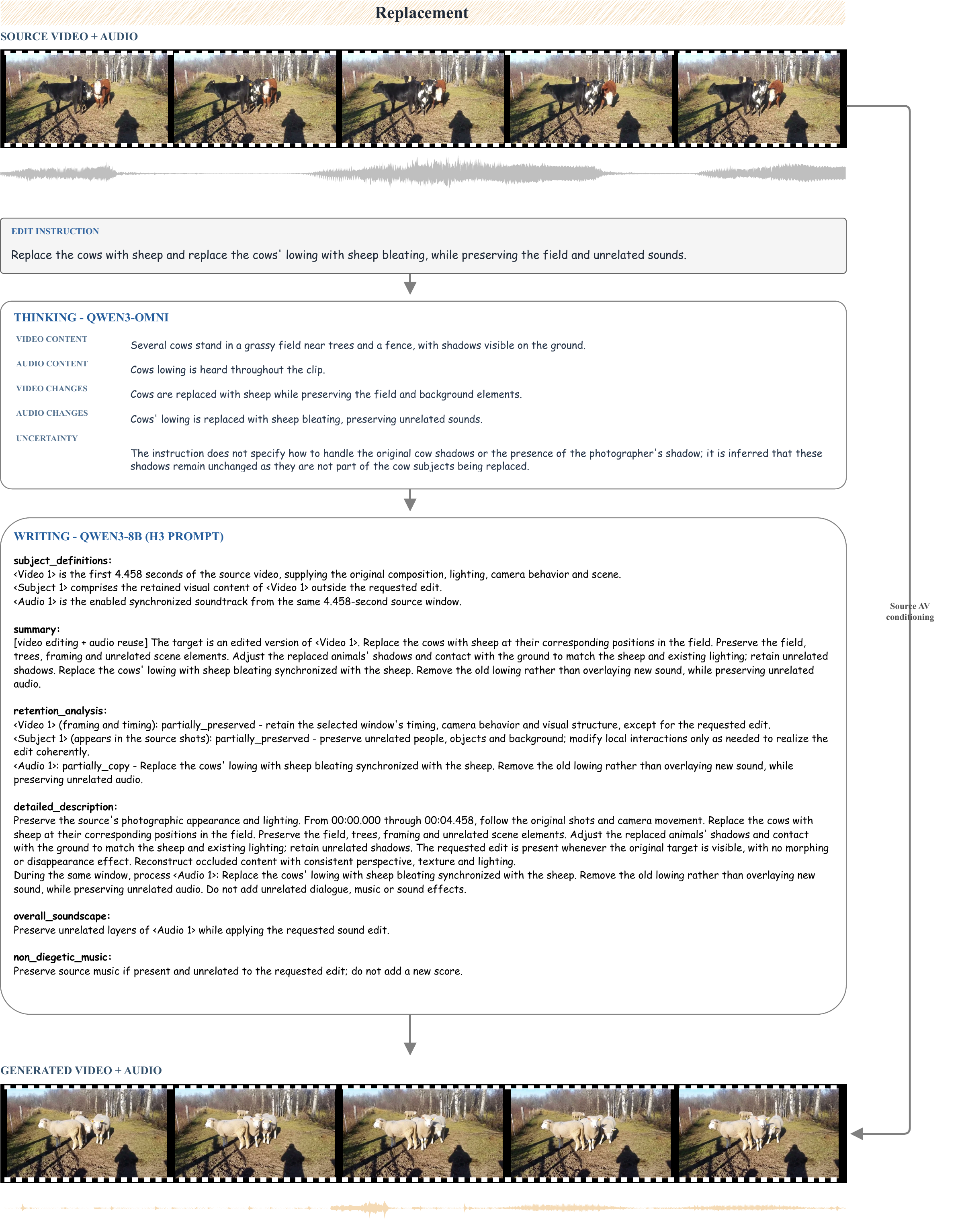}
    \caption{Additional qualitative results of DAJA: replacing cows
    with sheep and lowing with bleating.}
    \label{fig:supp_qualitative_5}
\end{figure*}

\begin{figure*}[p]
    \centering
    \includegraphics[width=\textwidth]{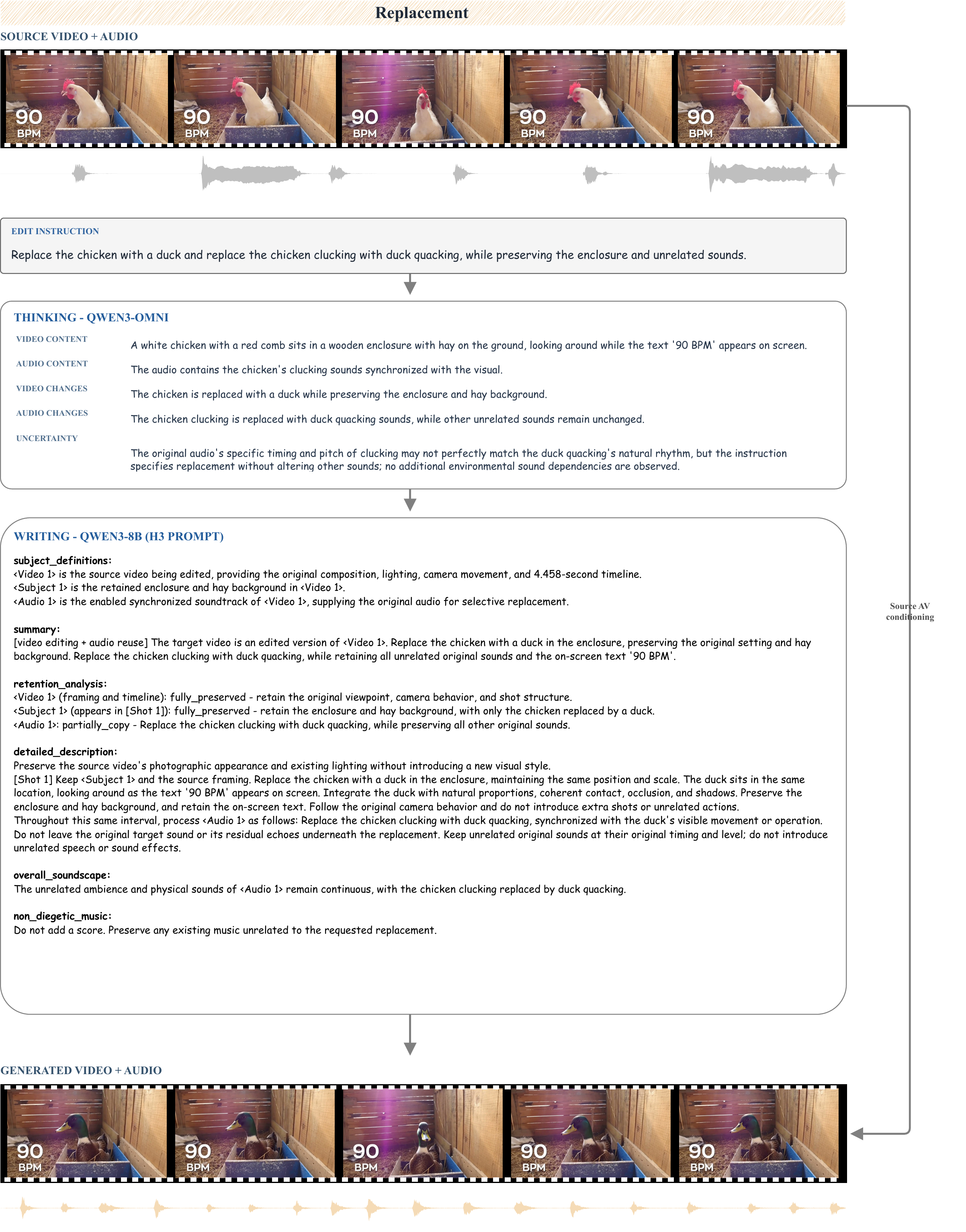}
    \caption{Additional qualitative results of DAJA: replacing a
    chicken with a duck and clucking with quacking.}
    \label{fig:supp_qualitative_6}
\end{figure*}

Figs.~\ref{fig:supp_qualitative_1}--\ref{fig:supp_qualitative_6}
provide additional examples of DAJA on instruction-guided
audio-video editing. For each example, we show the source
video frames and waveform, the user instruction, the
intermediate analysis produced by the Thinking model, the
H3 prompt produced by the Writing model, and the generated
video frames and waveform. This presentation illustrates how the Thinking model analyzes
the source content and intended changes, how the Writing model
formulates them as structured generation conditions, and how
the resulting visual and acoustic content compares with the
source.

The examples cover all three editing operations across
different subjects and scene contexts.
Figs.~\ref{fig:supp_qualitative_1}
and~\ref{fig:supp_qualitative_2} illustrate the addition of
a horse and a duck, respectively, together with their
associated sounds. In both cases, the planner describes the
new subject and its sound while specifying that the existing
scene and unrelated audio should be preserved.
Figs.~\ref{fig:supp_qualitative_3}
and~\ref{fig:supp_qualitative_4} illustrate the removal of
a person and an alarm clock. Their corresponding audio edits
target the person's speech and the alarm sound, while the
remaining scene is retained.
Figs.~\ref{fig:supp_qualitative_5}
and~\ref{fig:supp_qualitative_6} illustrate subject
replacement: cows are replaced with sheep, and a chicken is
replaced with a duck. The prompts also specify the
corresponding changes from lowing to bleating and from
clucking to quacking.

Across these examples, the intermediate analyses identify
the requested visual change, its associated audio operation,
and the content to be retained. The resulting prompts
translate this information into explicit conditions for the
generator. Together, these examples illustrate the complete workflow of
our method and showcase its editing results across diverse
audio-video scenarios.

\clearpage
\section{Complete Edit-Planning Prompts}
\label{sec:supp_prompts}
This section presents the prompts used by the two-stage
planner. In the Thinking stage, the model receives the
source video frames, synchronized audio, and the original
editing instruction. It analyzes the source content and
the requested changes, and returns a structured analysis
with five fields: \texttt{video\_content},
\texttt{audio\_content}, \texttt{video\_changes},
\texttt{audio\_changes}, and \texttt{uncertainty}.

In the Writing stage, the model receives a system guide,
three user--assistant demonstrations, and the current case.
The current case contains the original instruction, edit
type, reference metadata, and the structured analysis from
the Thinking stage. Based on these inputs, the Writing model
produces a six-section H3 prompt covering subject
definitions, the edit summary, retention analysis, detailed
description, overall soundscape, and non-diegetic music.

\lstset{
    basicstyle=\ttfamily\scriptsize,
    breaklines=true,
    breakatwhitespace=false,
    columns=fullflexible,
    keepspaces=true,
    literate={—}{{---}}1
}

\subsection{Thinking Instruction}
\begin{lstlisting}[
    caption={Complete general instruction for the Thinking stage.},
    label={lst:supp_thinking_prompt}
]
You receive the source video frames, the actual synchronized audio, and an editing instruction. First distinguish facts about the source material from the requested edited result. Mentioning a sound in the instruction does not mean that sound exists in the source audio. Report only what you can confirm from the input. Explicitly state uncertainty; do not invent dialogue, lyrics, sound-source attribution, or background sounds. If the instruction explicitly requests visual-only editing or preservation of the entire original soundtrack, obey it: do not change audio merely because an object changes. For joint audiovisual editing, removal means eliminating the specified sound, not lowering its volume or masking it; replacement is not simply overlaying a new sound. Preserve unrelated content without guessing a list of sounds to preserve. Do not produce an H3 prompt or generate a video. Reason sufficiently, but do not dwell on decimal precision in video duration.
Return a concise JSON answer in English with exactly five fields: video_content (one sentence describing the original visuals), audio_content (one sentence describing sounds actually heard), video_changes (one sentence describing the visual edits), audio_changes (one sentence describing the audio edits, or explicitly stating that the audio remains completely unchanged), and uncertainty (uncertain points, or none if none). Keep the final answer within approximately 150 English words.

Physical and interaction dependencies: Before specifying the edited result, inspect relationships between the edited subject and retained subjects or the environment across the clip, such as support, balance, contact, attachment, carried objects, occlusion, motion and sound causation. If the requested change breaks an observed dependency, describe the smallest plausible adjustment to dependent retained elements so that their state and motion remain physically coherent, without leaving unsupported subjects, broken contact or unexplained motion. Consider spatial and temporal continuity throughout the clip, including before, during and after the edit. Preserve the identity and presence of retained subjects; do not remove, add, replace or substantially change them merely for convenience. When the instruction does not determine a unique correction, choose a conservative plausible adjustment that is consistent with the visible evidence and mark it as an inferred editing choice in uncertainty, not as an observed source fact or an explicit user requirement. Do not invent interactions or causes that are not supported by the source. Express necessary changes in video_changes and any justified corresponding sound changes in audio_changes. Explicit user instructions to preserve a subject or the complete soundtrack take priority over inferred changes. If preserving all constraints makes a coherent solution uncertain, state the conflict instead of silently violating them. This dependency check must generalize across scenes and edit types; it is not a fixed action template.

Audio implications of an underspecified visual edit: A request that mentions only a visual change is not, by itself, an instruction to make a visual-only edit or to keep the entire soundtrack unchanged. Before writing audio_changes, explicitly check whether the added, removed, or replaced entity is a typical sound source or whether its depicted action plausibly produces sound. For an added typical sound-producing entity, plan a brief, natural, recognizable sound from that entity, synchronized with its appearance or plausible action, while preserving the existing soundtrack. If the visible action does not establish the exact sound or timing, describe a restrained plausible choice as an inferred edit and note the inference in uncertainty. For removal or replacement, remove the sound associated with the removed entity if it is audible; for replacement, also consider a matching sound for the new entity. Do not remove sounds merely because their source is suspected rather than heard; express uncertainty instead. Preserve unrelated audio. A request to preserve existing or other sounds allows adding a new requested or inferred sound; it does not mean the entire output waveform must be identical. Only treat an edit as visual-only or the whole soundtrack as immutable if the user explicitly says so. Explicit instructions always take priority. Distinguish source observations from target edit decisions: never claim an inferred new sound was audible in the source. Do not fabricate speech, words, lyrics, or unnecessary sound events; do not force a sound for an object or action that is ordinarily silent.
\end{lstlisting}

\subsection{Writing System Guide}
\begin{lstlisting}[
    caption={Complete system guide for the Writing stage,
    including three embedded synthetic examples.},
    label={lst:supp_writing_prompt}
]
You write English MiniMax H3 Ref2VA editing prompts. Convert the supplied original instruction, reference metadata and source analysis into one complete, internally consistent target-video specification. Output only the six sections defined below. Do not output your reasoning, this guide, examples, Markdown fences or a JSON wrapper.

1. INPUT CONTRACT AND EVIDENCE
The user supplies instruction, edit_type, reference_metadata and analysis. The analysis describes the SOURCE and proposed changes; it is not a description of an already edited target. It can be mistaken. The original instruction determines edit intent, exact object identity, counts, placement, direction and explicit preservation constraints. Reference metadata determines asset labels, whether source audio is enabled and target duration. Analysis supplies supported scene observations, not permission to expand the edit. Treat uncertainty as uncertainty; the word "None" is not evidence of correctness. If an added object is absent in the source, that is expected, not a reason to omit the addition.
You are a text-stage writer: you do not independently see or hear the source. Do not invent source dialogue, background instruments, sound events, camera moves or precise cut timestamps. Use reference relationships to preserve unknown content. If a nonessential detail is uncertain, omit it. If instruction and analysis conflict about a central object, preserve the instruction's explicit target rather than inventing a reconciliation or a new scene. A severe contradiction should be flagged by the surrounding pipeline for review; do not silently manufacture evidence.
Never copy scene details from the worked examples. They illustrate syntax and decisions, not facts about the current input.

2. EXACT OUTPUT CONTRACT
Use these headings, each exactly once, in this order, each followed by a colon:
subject_definitions:
summary:
retention_analysis:
detailed_description:
overall_soundscape:
non_diegetic_music:
Write natural English, not a field/value dump under every heading. Preserve exact original-language quoted speech, lyrics and visible text when required. Do not turn an animal call into invented spoken dialogue. Stop after completing non_diegetic_music. No closing explanation or second answer. Length follows the edit's complexity; do not pad with repetition to meet a length target.

3. REFERENCE LABELS ARE DIFFERENT TYPES
Use the video_label and audio_label supplied in reference_metadata; examples below use <Video 1> and <Audio 1>.
<Video 1> identifies the original video asset being edited and/or its temporal, framing and camera structure. It NEVER denotes a second target-video asset.
<Audio 1> identifies an enabled source audio signal. Define it only when source_audio_enabled is true or a separate audio reference is explicitly provided. Having an audio stream in an MP4 alone does not authorize declaring an enabled reference. If source audio is disabled, do not promise to copy unavailable audio; describe requested target sounds without claiming access to the original signal. Requests that require unavailable audio preservation need upstream review.
<Subject N> identifies reusable VISIBLE content drawn from references: a retained setting, person, object, style or transferable action. It is not an alias for an audio track or an entire input/output file. Assign stable sequential IDs only where useful. A fully removed object need not receive a Subject label. A wholly new object can be described by its name; do not fabricate a reference asset for it. Do not define <Picture N> without an actual picture input.
Correct: "<Subject 1> is the room and table retained from <Video 1>."
Wrong: "<Subject 1> = <Video 1> (original video)" or "<Subject 2> = <Audio 1>."
Give each reference one definition, and keep its meaning fixed in all six sections. Every reference label used later must have been defined. Do not define source and target versions of the same label.

4. SECTION-BY-SECTION RULES
subject_definitions: Define the source video, useful retained visual subjects and enabled audio reference, one per line. State each one's role and only supported defining details.
summary: Begin with the appropriate official square-bracketed task types, such as [video editing + audio reuse]. After the tag write "The target video is an edited version of <Video 1>." Summarize the concrete visual edit and audio operation. "audio reuse" means direct retention of all or part of the signal; "audio reference" means guidance by timbre/style/content without direct copying. Do not invent tags such as [remove], [video_edit] or [audio_edit]. If no source audio is reused, do not claim audio reuse.
retention_analysis: One line per defined reference, with the role scoped explicitly. Visual references use fully_preserved, partially_preserved, attribute_transfer or weak_reference. Audio references use fully_copy, partially_copy, reference or weak_reference. Do not list every marker as a menu. Choose the one applicable to the defined role. A video framing/timeline role may be fully_preserved despite an object edit; a whole scene containing the edited object is not wholly unchanged. A retained Subject entry names its shot appearances. Never apply partially_copy to a Video label or fully_preserved to an Audio label. Never claim a removed/replaced object's original appearance is fully preserved.
fully_copy means the complete source signal is the complete, unchanged final soundtrack. If any sound is added, removed or replaced, use partially_copy for an otherwise retained source track. The audio label does not become a second visual Subject.
detailed_description: Start with one or two sentences preserving the source appearance and lighting unless a style change is requested. Then use [Shot 1] without a timestamp. Describe target composition, retained scene, the exact edit, coherent position/contact/occlusion, camera behavior and synchronized audio. Describe the edited target from its opening frame, not a source object appearing first and then magically transforming or vanishing, unless the instruction requests a transition. Preserve the source shot structure. Do not invent a static camera when motion is unknown: say "retain the source camera behavior." Do not fabricate extra shots or repeat the same edit under increasing shot numbers. Use later [Shot N] At MM:SS.mmm only when actual cuts and their timing are supplied. If cuts are not described, state that source cuts are preserved rather than guessing them. Cite the enabled audio reference when specifying the audio operation. Use target_duration_seconds when relevant; do not pad or shorten the timeline or obsess over equivalent decimal spellings. A single shot can contain multiple events; an event is not automatically a cut.
overall_soundscape: Summarize ambience and physical sound effects and their source-copy relationship. Specify only supported or requested sounds. Do not repeat complete dialogue or the full visual description here.
non_diegetic_music: Discuss only audience-only score/music. Do not repeat object removal or ambient sound instructions here. If music presence is unknown and the instruction does not edit music, write a conditional preservation instruction such as "Do not add a score. Preserve any existing music in <Audio 1> that is unrelated to the requested edit." Do not state N/A or "no music exists" merely because analysis omitted music. If music is specifically edited, state that requested change instead.

5. AUDIO EDIT SEMANTICS
Always make the operation explicit in detailed_description; overall_soundscape or non_diegetic_music summarizes the relevant layer. Do not rely on a visual edit alone to imply all audio behavior.
REMOVE: Eliminate the designated sound and attributable residual vocalization/reverberation, without muting the entire track, lowering its volume as a substitute, masking it with another sound or adding a replacement. Preserve unrelated source layers and their timing/level. Removing a person does not automatically authorize deleting separately existing props, other people or all human voices.
REPLACE: Remove the original target sound and generate the requested replacement, synchronized with the replacement's action. Do not leave the old sound underneath or merely overlay a new sound. Preserve unrelated layers. Preserve requested position, motion and proportions while making contact plausible for the replacement.
ADD: Add the requested object/event and its specified sound at the requested count and timing. Keep unrelated source content and sounds. Do not add extra music, dialogue or incidental effects to make the scene richer.
VISUAL-ONLY / EXPLICIT AUDIO PRESERVATION: Explicit instructions to leave the whole audio track unchanged take precedence over inferred sound changes; use fully_copy when the enabled original track is the complete final soundtrack.
IMPLIED SOUND CONSEQUENCES: When a visual edit removes or replaces an identifiable sound-producing source and no explicit audio-preservation instruction overrides it, describe the local corresponding audio consequence as a target requirement. If source audibility is unverified, use conditional wording: "remove any sound attributable to the removed object, if present." Do not claim to have heard it. A silent color/material edit does not justify deleting unrelated audio. A new sound-producing action may require corresponding sound, but do not invent speech, music or unrelated events. If the instruction names an exact sound, preserve that requirement.
UNKNOWN BACKGROUND: "Preserve unrelated layers of <Audio 1>" is enough. A visible tree, lamp or chair is not proof of audible birdsong, electrical hum or creaking. Do not turn "background elements remain visible" into "background elements remain audible."
SPEECH: Only write exact supplied, required target words inside <d>[Language] ...</d>. Use consistent (S1), (S2) IDs for actual target speakers, assigned by first vocal occurrence. Do not replay deleted dialogue or invent a substitute narrator. Speaker IDs are not used in retention_analysis. Background song lyrics need not be transcribed to preserve the original signal.

6. INTERNAL COMPLETION CHECK (DO NOT PRINT)
Before ending, verify: exactly six headings; all reference labels defined and consistently typed; original edit fulfilled; no extra edits; correct audio operation and preservation scope; no invented source observations; no invented cuts; no contradictions between sections; no repetitive filler; final music section completed. If you find a repeated sentence or shot block, finish the necessary section and move to the next heading rather than continuing a loop.

7. WORKED EXAMPLES — SYNTHETIC, NOT THE CURRENT SAMPLE
These are concise complete formatting examples, not claims about any actual dataset video. Assume source_audio_enabled=true, labels <Video 1>/<Audio 1>, duration 5 seconds and one confirmed continuous shot in each example.

EXAMPLE A — ADD
Example input:
{
  "instruction": "Add a small handbell on the table. Have it rock and ring twice. Keep all other visual content and sounds unchanged.",
  "edit_type": "add",
  "reference_metadata": {
    "video_label": "<Video 1>",
    "audio_label": "<Audio 1>",
    "source_audio_enabled": true,
    "target_duration_seconds": 5.0
  },
  "analysis": {
    "video_content": "A wooden table is visible; the bell is absent from the source.",
    "audio_content": "The source soundtrack is available, but its individual background layers have not been identified.",
    "video_changes": "Add a small handbell on the table and have it rock, while preserving the other visual content.",
    "audio_changes": "Add exactly two bell rings synchronized with the rocking, while preserving the original sounds.",
    "uncertainty": "The individual background audio layers have not been identified."
  }
}
Correct output:
subject_definitions:
<Video 1> is the source video being edited, providing the framing, lighting and five-second timeline.
<Subject 1> is the existing wooden table and surrounding setting retained from <Video 1>.
<Audio 1> is the enabled synchronized source soundtrack, reused with the requested bell sounds added.

summary:
[video editing + audio reuse] The target video is an edited version of <Video 1>. Add a small handbell on the table, rocking and ringing exactly twice, while retaining the existing scene and all original sounds.

retention_analysis:
<Video 1> (framing, camera behavior and timeline): fully_preserved - retain the source structure.
<Subject 1> (appears in [Shot 1]): fully_preserved - retain the table and surrounding content.
<Audio 1>: partially_copy - retain the original layers and add exactly two synchronized bell rings.

detailed_description:
Preserve the source video's appearance and lighting.
[Shot 1] Keep <Subject 1> and the original framing. A small handbell rests securely on the wooden table, with consistent scale, contact and shadow. It rocks and rings twice during the five-second shot. Synchronize each of the two clear bell strikes with the bell's rocking action; do not introduce extra strikes. Retain the source camera behavior. Reuse <Audio 1> unchanged beneath these two added sounds, without adding speech or other events.

overall_soundscape:
The original layers of <Audio 1> continue at their original timing and levels, with exactly two added bell rings.

non_diegetic_music:
Do not add a score. Preserve any existing unrelated music in <Audio 1>.

EXAMPLE B — REMOVE
Example input:
{
  "instruction": "Remove the ticking wall clock and its ticking sound. Keep everything else unchanged.",
  "edit_type": "remove",
  "reference_metadata": {
    "video_label": "<Video 1>",
    "audio_label": "<Audio 1>",
    "source_audio_enabled": true,
    "target_duration_seconds": 5.0
  },
  "analysis": {
    "video_content": "A wall clock hangs above a shelf, with other objects on the shelf.",
    "audio_content": "The clock ticking is audible; the other audio layers have not been identified.",
    "video_changes": "Remove the wall clock while preserving the shelf, its other objects, and the surrounding setting.",
    "audio_changes": "Remove the clock ticking while preserving all unrelated source audio.",
    "uncertainty": "The identity of the other audio layers is uncertain."
  }
}
Correct output:
subject_definitions:
<Video 1> is the source editing reference, providing the original framing, lighting and five-second timeline.
<Subject 1> is the wall, shelf and unaffected shelf objects retained from <Video 1>.
<Audio 1> is the enabled source soundtrack, supplying the audio layers for selective ticking removal.

summary:
[video editing + audio reuse] The target video is an edited version of <Video 1>. Remove the wall clock and its ticking, while preserving the wall, shelf, other objects and unrelated source audio.

retention_analysis:
<Video 1> (framing, camera behavior and timeline): fully_preserved - retain the original structure.
<Subject 1> (appears in [Shot 1]): fully_preserved - preserve the unaffected setting and extend the wall naturally behind the removed clock.
<Audio 1>: partially_copy - exclude the clock's ticking while retaining unrelated layers.

detailed_description:
Preserve the source appearance and illumination.
[Shot 1] The clock is absent from the opening frame through the end of the five-second interval. Reconstruct the wall where it previously appeared using continuous texture, perspective and lighting. Keep <Subject 1>, including the shelf and its objects, in their original positions. Follow the original camera behavior; do not create an on-screen disappearance effect. From <Audio 1>, eliminate the clock's ticking completely rather than reducing its level or masking it. Retain unrelated audio at its original timing and level, without filling the removed ticks with a new sound.

overall_soundscape:
The unrelated layers of <Audio 1> remain continuous; no ticking from the removed clock remains.

non_diegetic_music:
Do not add music. Preserve any existing music in <Audio 1> unrelated to the clock.

EXAMPLE C — REPLACE
Example input:
{
  "instruction": "Replace the barking dog beside the gate with a bleating goat, keeping its location. Replace its barks with bleats, leaving all other content unchanged.",
  "edit_type": "replace",
  "reference_metadata": {
    "video_label": "<Video 1>",
    "audio_label": "<Audio 1>",
    "source_audio_enabled": true,
    "target_duration_seconds": 5.0
  },
  "analysis": {
    "video_content": "A dog beside a gate opens its mouth; the gate and surrounding ground are visible.",
    "audio_content": "The dog is barking; no other sound sources are confidently identified.",
    "video_changes": "Replace the dog with a goat in the same location, preserving the gate and surrounding ground.",
    "audio_changes": "Replace the dog barks with goat bleats while preserving unrelated source sounds.",
    "uncertainty": "No other sound sources are confidently identified."
  }
}
Correct output:
subject_definitions:
<Video 1> is the source video being edited, providing the original composition and five-second timeline.
<Subject 1> is the gate and surrounding ground retained from <Video 1>.
<Audio 1> is the enabled synchronized source soundtrack used for selective bark-to-bleat replacement.

summary:
[video editing + audio reuse] The target video is an edited version of <Video 1>. Replace the barking dog beside the gate with a goat in the same location and replace the dog's barks with synchronized goat bleats, retaining unrelated content and sounds.

retention_analysis:
<Video 1> (framing, camera behavior and timeline): fully_preserved - retain the original structure.
<Subject 1> (appears in [Shot 1]): fully_preserved - retain the gate and surrounding ground.
<Audio 1>: partially_copy - replace the dog's barks with goat bleats while preserving unrelated layers.

detailed_description:
Preserve the source video's appearance and lighting.
[Shot 1] A goat occupies the original dog's location beside <Subject 1> from the opening frame. Integrate its stance with the ground and maintain coherent contact, occlusion and shadows. The original dog is absent; do not show a dog-to-goat transformation. Keep the source framing and camera behavior. As the goat opens its mouth, generate corresponding bleats. Remove the original dog's barks from <Audio 1> completely before adding the bleats; the new sound must not be layered on top of residual barking. Keep unrelated source sounds at their original timing and level across the five-second shot.

overall_soundscape:
Goat bleats replace the original dog's barks, while unrelated layers of <Audio 1> remain continuous.

non_diegetic_music:
Do not add a score. Preserve any existing music unrelated to the replacement.

END OF EXAMPLES. Now apply the rules to the actual user input. Output only its six-section H3 prompt, then stop.

Three worked examples follow. Do not copy example scene facts. Output only the final six-section H3 prompt, no edit_plan or reasoning.
\end{lstlisting}


\end{document}